\documentclass[runningheads]{llncs}
\usepackage[T1]{fontenc}
\usepackage{graphicx}
\usepackage{amsmath,amssymb} % define this before the line numbering.

\usepackage{color}
\newcommand{\UrlFont}{\color{blue}\rmfamily}

\usepackage{subcaption}

\newcommand{\etal}{{\it et al.}}
\newcommand{\ie}{{\it i.e.}}
\newcommand{\eg}{{\it e.g.}}

\newcommand{\figref}[1]{Fig.~\ref{#1}}
\newcommand{\tabref}[1]{Tab.~\ref{#1}}

\newcommand{\secref}[1]{Sec.~\ref{#1}}
\renewcommand{\eqref}[1]{Eq.~\ref{#1}}

\newcommand\blfootnote[1]{%
  \begingroup
  \renewcommand\thefootnote{}\footnote{#1}%
  \addtocounter{footnote}{-1}%
  \endgroup
}

\begin{document}
\title{NoiseTransfer: Image Noise Generation with Contrastive Embeddings}
\titlerunning{NoiseTransfer}
\author{Seunghwan Lee \and
Tae Hyun Kim}
\authorrunning{Lee and Kim.}
\institute{Dept. of Computer Science, Hanyang University, Seoul, Korea
\email{\{seunghwanlee,taehyunkim\}@hanyang.ac.kr}}
\maketitle              % typeset the header of the contribution
%

%===========================================================
\begin{abstract}
Deep image denoising networks have achieved impressive success with the help of a considerably large number of synthetic train datasets.
However, real-world denoising is a still challenging problem due to the dissimilarity between distributions of real and synthetic noisy datasets.
Although several real-world noisy datasets have been presented, the number of train datasets (\ie, pairs of clean and real noisy images) is limited, and acquiring more real noise datasets is laborious and expensive.
To mitigate this problem, numerous attempts to simulate real noise models using generative models have been studied.
Nevertheless, previous works had to train multiple networks to handle multiple different noise distributions.
By contrast, we propose a new generative model that can synthesize noisy images with multiple different noise distributions.
Specifically, we adopt recent contrastive learning to learn distinguishable latent features of the noise.
Moreover, our model can generate new noisy images by transferring the noise characteristics solely from a single reference noisy image.
We demonstrate the accuracy and the effectiveness of our noise model for both known and unknown noise removal.\blfootnote{Code is available at \UrlFont{https://github.com/shlee0/NoiseTransfer}}

\end{abstract}
%
%
%

%===========================================================

\begin{figure*}[hbt!]
  \centering
  \includegraphics[width=\textwidth]{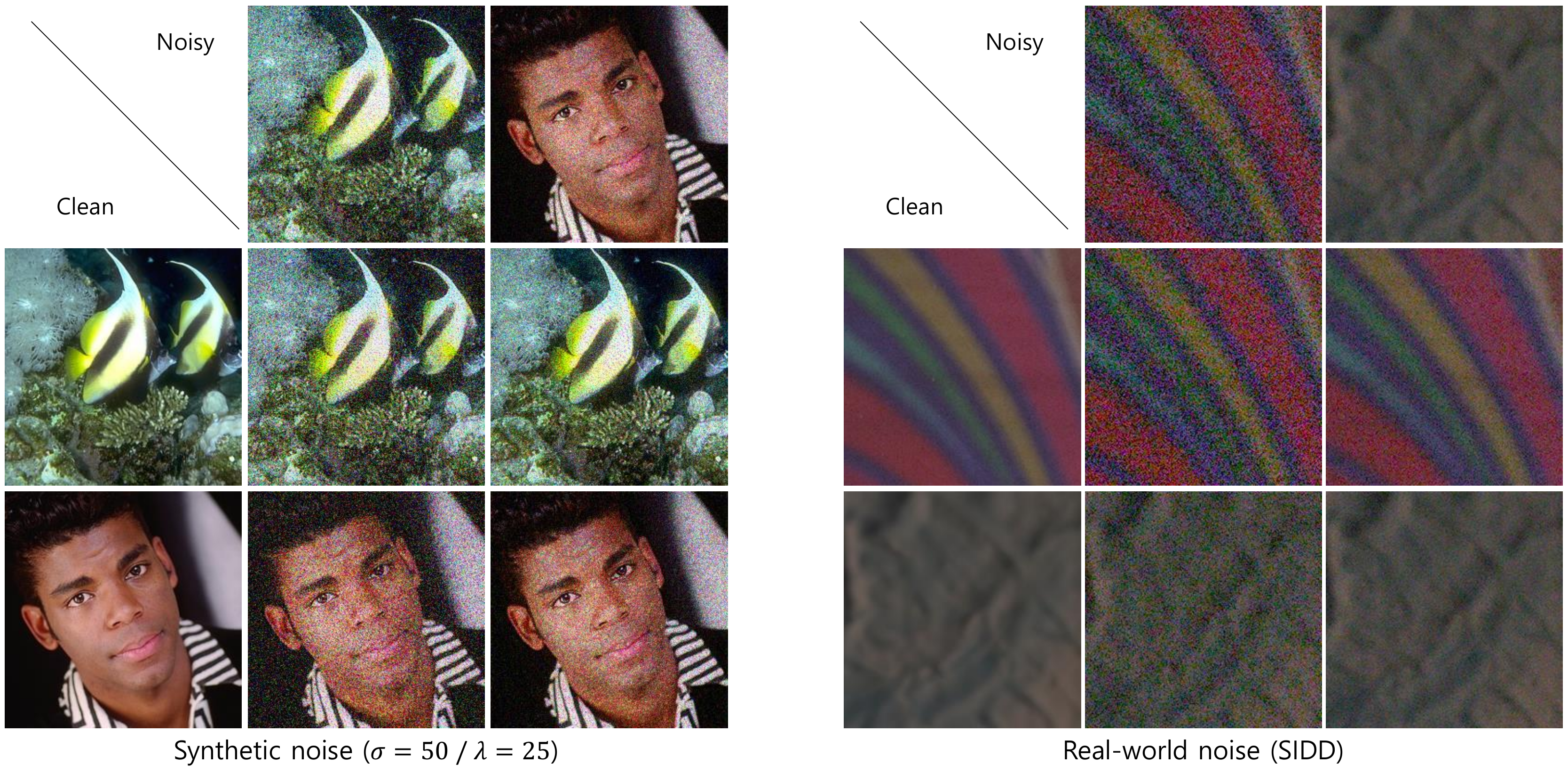}
  \caption{Examples of generated noisy images. Our proposed model can generate a noisy version of a clean image by transferring noise information in the reference noisy image. \textbf{(Left)} Synthetic noise (\ie, Gaussian noise ($\sigma$ = 50) (2nd column) and Poisson noise ($\lambda$ = 25) (3rd column)) generation results. \textbf{(Right)} Real-world noise generation results from the Smartphone Image Denoising Dataset (SIDD) \cite{SIDD}. Noisy images generated in an unpaired manner can have different noise levels that do not exist in the original dataset.}
  \label{fig_teaser}
\end{figure*}

\section{Introduction}

% Intro-1: statistical methods로는 real-world denoising 한계가 있음
Noise is a common artifact in imaging systems; thus, accurately modeling the noise is an important task for many image-processing and computer vision applications.
To remove the noise in an image, several statistical noise models have been adopted in the literature.
The most simple, widely used noise models are additive white Gaussian noise and Poisson noise.
However, in a real-world scenario, image noise does not follow Gaussian and Poisson distributions~\cite{DND,SIDD}, and  these simple statistical noise models cannot accurately capture the noise characteristic of real noise which includes signal-dependent and signal-independent components.
Moreover, developing a noise model that can simulate complex real-world noise process is very difficult because complicated processing steps of an imaging pipeline include various noise models, such as photon noise, read noise, and spatially correlated noise.
Conventional denoising networks that show promising results in removing noise from known distribution (\eg, Gaussian) frequently fail in dealing with real noise from an unknown distribution due to these limitations.

% Intro-2: real noise simulation methods
Collecting real-world datasets that include pairs of clean and real-noisy images can solve these problems.
However, the noise distributions of conventional cameras are different from one another, so we need to acquire a large amount of labelled real-world dataset, which is very time-consuming.
This problem stimulates the need for synthetic, but realistic noise generation system to avoid taking pairs of clean and noisy pictures.
Recently, several generative adversarial network (GAN)-based noise models have been proposed to model the complex real-world noise in a data-driven manner better.
Since Chen \etal~\cite{chen2018image} has proposed a generative model to synthesize zero-mean noise, recent models~\cite{grdn,noiseflow,UIDNet,CA-NoiseGAN,DANet,syntGAN,C2N,PNGAN} made many attempts to generate signal-dependent noise by considering a clean image as a conditional input.

% Intro-3: existing 모델들의 한계: noise sample uncertainty
Despite this encouraging progress, there are still some steps to move forward for image noise generation.
Typically, generative models have difficulty in controlling the specific type of noise during synthesizing.
In other words, which types of noise will be realized is not predictable at the inference if generator is trained with a large of noise distributions.
In addition, this randomness increases if the training dataset includes several different noise types.
A naïve, straightforward solution would be to train multiple generators independently to handle multiple noise models.
Alternatively, image metadata such as camera-ISO and the raw Bayer pattern can be utilized to avoid this hassle.
However, this external data is not always available (\eg, images from unknown resources).

% Intro-4: our method & contribution
In this work, we propose a novel generative noise model, which can allow multiple different types of noise models.
We transfer the noise characteristics within a given reference noisy image to corrupt freely available clean images, and we synthesize new noisy images in this manner.
Moreover, our model requires only the noisy image itself without demanding any external information (\eg, metadata).
Specifically, we train our discriminator to distinguish the distribution of each noise from the others in a self-supervised manner by adopting a contrastive learning.
Then, our generator learns to synthesize a new noisy image using the noise information extracted from the discriminator.
With this strategy, we can perform noise generation with paired or unpaired images, and \figref{fig_teaser} presents some examples.
We demonstrate that our generative noise model can handle a wide range of noise distributions, and the conventional denoising networks trained with our newly synthesized noisy images can remove the real noise much better than existing generative noise models.
The main contributions of our work are summarized as follows:
\begin{itemize}
\item We propose a novel generative noise model that can handle diverse noise distributions with a single noise generator without additional meta information.
\item Our model exploits the representation power of the contrastive learning. To the best of our knowledge, our model is the first approach which utilizes contrastive noise embedding to control the type of noise to be generated.
\item Extensive experiments demonstrate that our model achieves state-of-the-art performance in noise generation and is applicable for image denoising.
\end{itemize}
%Do not use any additional Latex macros.

\section{Related Work}

% 1.1. contrastive learning
\subsection{Contrastive learning}
The contrastive learning mechanism introduced by~\cite{hadsell2006dimensionality} learns similar/dissimilar representations in a self-supervised manner from positive/negative pairs.
In the works of instance discrimination~\cite{NEURIPS2019_ddf35421,Ye_2019_CVPR}, a query and a key form a positive pair if they originate from the same image and form a negative pair if otherwise.
It is known that more negative samples can yield better representation ability, and a large number of negative samples can be maintained in a batch~\cite{SimCLR} or dynamic dictionary updated by a momentum-based key encoder~\cite{MoCo}.

% 1.2. contrastive learning + GAN
After the contrastive learning has shown powerful representation ability in several downstream tasks, it has been integrated with the GAN framework as an auxiliary task.
For instance, contrastive learning could relieve forgetting problem of discriminator~\cite{chen2019self,InfoMaxGAN}, and improve image translation quality by maximizing mutual information of corresponding patches in different domains~\cite{CUT,DCLGAN}.
ContraGAN~\cite{ContraGAN} improved image generation quality by incorporating data-to-data relations as well as data-to-class relations into discriminator.
ContraD~\cite{ContraD} empirically showed that training the GAN discriminator jointly with the augmentation techniques used in the literature of contrastive learning benefits the task of the discriminator.
% In DASR~\cite{DASR}, positive examples of low-resolution images with the same degradation are constructed and the representation of content-invariant degradation was learned.
Moreover, contrastive learning can learn content-invariant degradation representation by constructing image pairs with the same degradation as positive examples.
Recently, DASR~\cite{DASR} and AirNet~\cite{AirNet} utilized learned degradation representation for image restoration.
% However, image synthesis conditioned on representation learned through contrastive learning has not yet been studied.
% Our work is differentiated from previous works in that we directly exploits the contrastive embeddings as a conditional input for image noise generation.
Different from previous works, our work studies image noise synthesis conditional on degradation representation learned through contrastive learning.

% 2. generative noise model
\subsection{Generative noise model}
To address the limitations of simple synthetic noise models, considerable effort has been devoted to numerous generative noise models to synthesize complex noise for the real-world image denoising problem.
Particularly, recent generative noise models yield signal-dependent noise given a clean image.
Some approaches require metadata (\eg, smartphone code, ISO level, and shutter speed) as an additional input to generate noise from a specific distribution~\cite{grdn,noiseflow,CA-NoiseGAN}.
However, these approaches assume that the metadata is available, which might not be common in the real scenario (\eg, internet images and pictures),
and the use of this additional information limits the usage of the generative noise model in practice.
Unlike existing generative models, our model extracts noise representation from an input noisy image itself without relying on the metadata, and thus allows us to use any noisy image as a reference.
Then, our generator synthesizes new noisy images based on noise information of the reference noisy image, such that we can easily predict which type of noise will be realized.

\begin{figure*}[hbt!]
  \centering
  \includegraphics[width=1.0\linewidth]{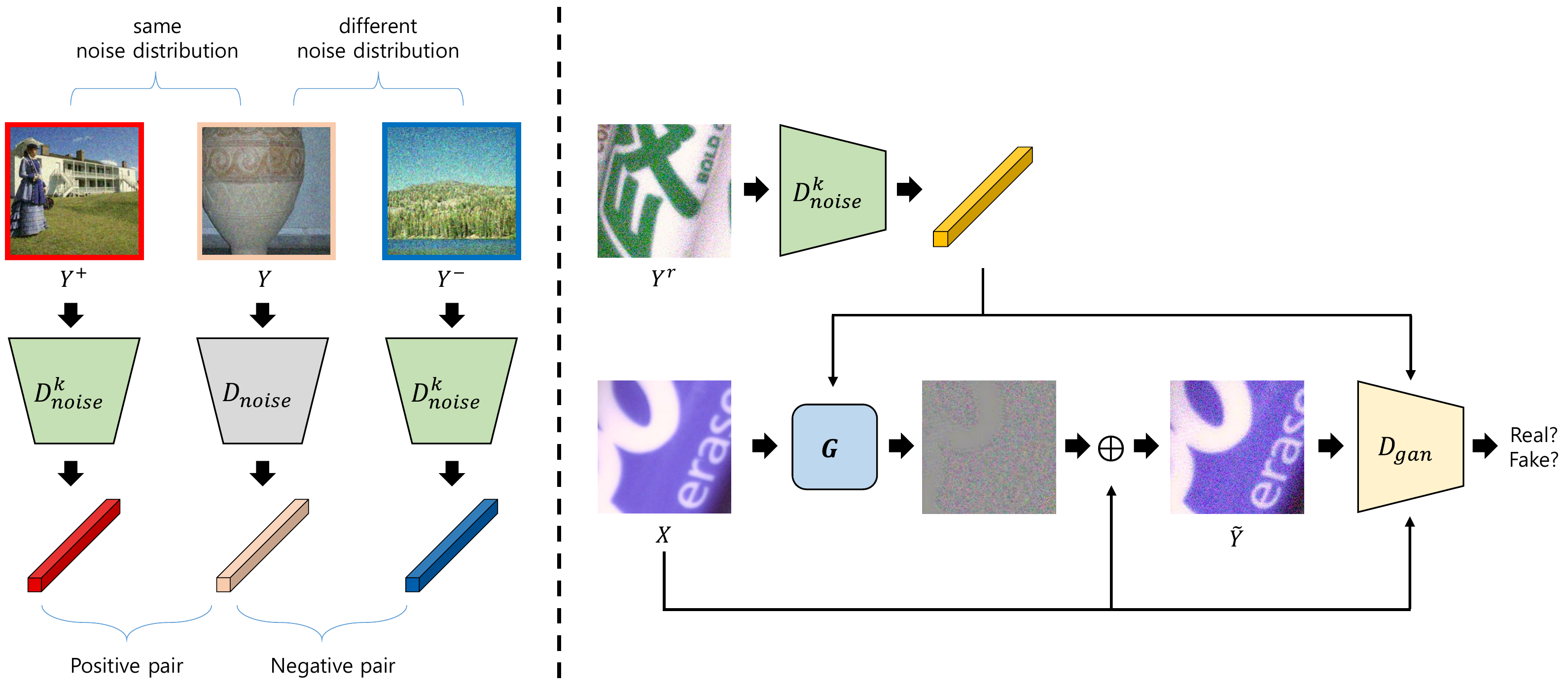}
  \caption{Our discriminator consists of two branches with shared intermediate convolutional modules for each forward operation denoted by $D_{noise}$ and $D_{gan}$ respectively. \textbf{(Left)} An illustration of our noise representation learning scheme. Two noisy images sampled from the same noise distribution form a positive and, in different cases, a negative. \textbf{(Right)} Overall flow of the proposed  NoiseTransfer. Our noise generator takes a clean image $X$ and noise embeddings $D_{noise}^k(Y^r)$ where $Y^r$ is a reference noisy image.}
  \label{fig_overview}
\end{figure*}

\section{Proposed Method: NoiseTransfer}\label{sec_method}
Our generative noise model synthesizes new noisy images by transferring the noise of a reference noisy image to other clean images.
Specifically, our discriminator takes a single reference noisy image as an input and outputs noise embeddings that represent noise characteristics of the reference noisy image.
Then, our generator synthesizes new noisy images by corrupting clean images available for free using the given noise embeddings that we dub NoiseTransfer.
\figref{fig_overview} depicts the overview of the proposed NoiseTransfer scheme.

\subsection{Noise discrimination with contrastive learning}
Capturing different characteristics for different noises is essential to keep the noise information distinct.
Therefore, we train our discriminator through contrastive learning to learn distinguishable noise embeddings of each noise, and we follow MoCo~\cite{MoCo} framework: dynamic dictionary holding a large number of negative samples and momentum-based key network.
Then, a form of a contrastive loss function, called InfoNCE~\cite{infonce}, can be written with cosine similarity $s(u,v)=u \cdot v / \Vert u \Vert_{2} \Vert v \Vert_{2}$ for encoded embeddings $u$ and $v$ as follows:
% equ, infonce loss
\begin{equation}
\begin{split}
    L_{\mathrm{Con}}&(q,k^+,Q) = \\&-\log \frac{\exp(s(q,k^+) / \tau)}{\exp(s(q,k^+) / \tau) + \sum\limits_{k^- \in Q} \exp(s(q,k^-) / \tau)},
\end{split}
\label{equ_contrastive_loss}
\end{equation}
where $q$, $k^+$, and $k^-$ denote the embeddings of a query, positive key, and negative key, respectively; $Q$ denotes a queue containing negative keys; and $\tau$ is a temperature hyperparameter.
\eqref{equ_contrastive_loss} pulls embeddings of the $q$ close to those of the $k^+$ and pushes them apart from those of the $k^-$.
% Note that we encodes the keys with momentum-updated key discriminator $D^k$.
% The $k^+$ and $k^-$ are encoded by momentum-based key network $D^k$~\cite{MoCo}.

In our work, as shown in~\figref{fig_overview} (Left), we construct a positive pair of noisy images if they are sampled from the same noise distribution and a negative pair, otherwise.
Then, the contrastive loss for noise discrimination can be formulated as follows:
\begin{equation}
    L_{noise}^D = \mathbb{E} [L_{\mathrm{Con}}(D_{noise}(Y), D_{noise}^k(Y^+), Q)],
\label{equ_loss_noise}
\end{equation}
where $Y^+$ denotes a noisy image that has the same noise distribution to that of another noisy image $Y$.
Note that we encodes the keys ($k^+$ and $k^-$) with momentum-based key network $D_{noise}^k$.
We assume that embeddings in $Q$ are from noisy images whose noise distributions are different from that of $Y$.
\eqref{equ_loss_noise} encourages our discriminator to learn distinguishable noise representation for each different noise.

Our final goal is to synthesize a new noisy image  $\tilde{Y}$ through a generator, which has the same noise distribution as the real one $Y$. 
Thus, we derive another contrastive loss for the generator as follows:
\begin{equation}
    L_{noise}^G = \mathbb{E} [L_{\mathrm{Con}}(D_{noise}(\tilde{Y}), D_{noise}^k(Y^+), Q)].
\label{equ_loss_G_noise}
\end{equation}
Note that, in \eqref{equ_loss_G_noise}, our generated noisy image $\tilde{Y}$ is encoded as a query.
Moreover, we adopt a feature matching loss~\cite{NIPS2016_8a3363ab} to stabilize training as follows:
\begin{equation}
    L_{noise}^{FM} = \Vert m_{noise}(Y) - m_{noise}(\tilde{Y}) \Vert_1,
\label{equ_loss_G_noise_fm}
\end{equation}
where $m_{noise}(\cdot)$ denotes the intermediate feature maps before pooling operation in the $D_{noise}$ (please refer to the supplement for details).

\subsection{Noise generation with contrastive embeddings}
Given a clean image $X$ and a reference noisy image $Y^{r}$, our generator learns to synthesize a new noisy image $\tilde{Y}$ which is a noisy version of $X$ and has the same noise distribution as $Y^{r}$.
The generation process is described in \figref{fig_overview} (Right).
Reference noisy image $Y^{r}$ is encoded by $D_{noise}^k$, and noise embeddings $D_{noise}^k(Y^{r})$ that contain noise representation of the $Y^{r}$ are fed to our generator.
This approach enables our model to handle a wide range of noise distributions with a single generator.
To generate realistic noisy images, our model performs adversarial training.
The adversarial losses~\cite{GAN} for our model are defined as follows:
\begin{equation}
    \begin{split}
    & L_{gan}^D = - \mathbb{E}[log(D_{gan}(R))] - \mathbb{E}[log(1 - D_{gan}(F)] \\
    & L_{gan}^G = - \mathbb{E}[log(D_{gan}(F))],
    \end{split}
\label{equ_loss_gan}
\end{equation}
where $R$ denotes a set of $X, D_{noise}^k(Y^{r})$, and $Y$, whereas $F$ includes $\tilde{Y}$ instead of $Y$.
Our generator synthesizes noisy images with different kinds of noise distribution based on the $D_{noise}^k(Y^{r})$, even with the same clean image $X$. Thus, our discriminator distinguishes whether the input noisy image is real or fake considering the $X$ and $D_{noise}^k(Y^{r})$.
Similar to \eqref{equ_loss_G_noise_fm}, we adopt feature matching loss for stable adversarial training as follows:
\begin{equation}
    L_{gan}^{FM} = \Vert m_{gan}(Y) - m_{gan}(\tilde{Y}) \Vert_1,
\label{equ_loss_G_dis_fm}
\end{equation}
where $m_{gan}(\cdot)$ denotes feature maps before the last convolution layer in the $D_{gan}$.
Finally, we utilize $L_1$ reconstruction loss $L_{recon} = \Vert \textit{GF}(Y) - \textit{GF}(\tilde{Y}) \Vert_1$ with the Gaussian filter \textit{GF} as used in \cite{DANet} to enforce statistical features of noise distribution.
Then, we define the final objective functions for our model as follows: 
\begin{equation}
    \begin{split}
    & L_{\mathrm{D}} = L_{noise}^D + L_{gan}^D \\
    &   \begin{split}
        L_{\mathrm{G}} = 
        & L_{noise}^G + L_{gan}^G + \\
        & \lambda_{noise}^{FM} L_{noise}^{FM} + \lambda_{gan}^{FM} L_{gan}^{FM} + \lambda_{recon} L_{recon},
        \end{split}
    \end{split}
\label{equ_loss_final}
\end{equation}
where $\lambda_{noise}^{FM}, \lambda_{gan}^{FM}$, and $\lambda_{recon}$ control the weights of the associated terms.

\begin{figure*}[hbt!]
  \centering
  \includegraphics[width=1.0\linewidth]{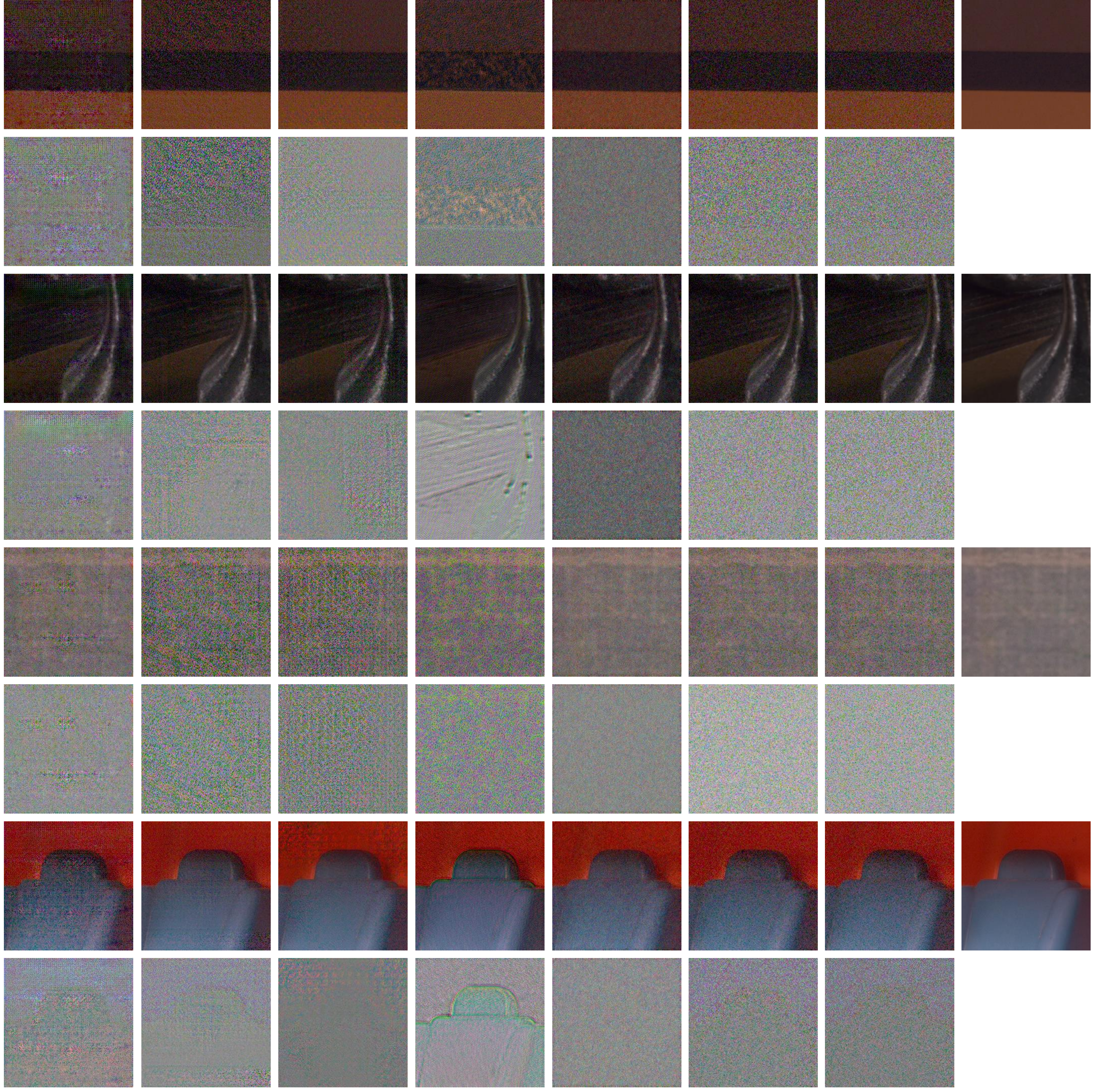}
  \caption{Visual results of noise generation on the SIDD validation set. The corresponding noise is displayed below for each noisy image. \textbf{Left to Right}: CA-NoiseGAN~\cite{CA-NoiseGAN}, DANet~\cite{DANet}, GDANet~\cite{DANet}, NoiseGAN~\cite{syntGAN}, C2N~\cite{C2N}, NoiseTransfer (Ours), Noisy, Clean.}
  \label{fig_sidd_generation}
\end{figure*}

\begin{figure*}[hbt!]
  \centering
  \includegraphics[width=1.0\linewidth]{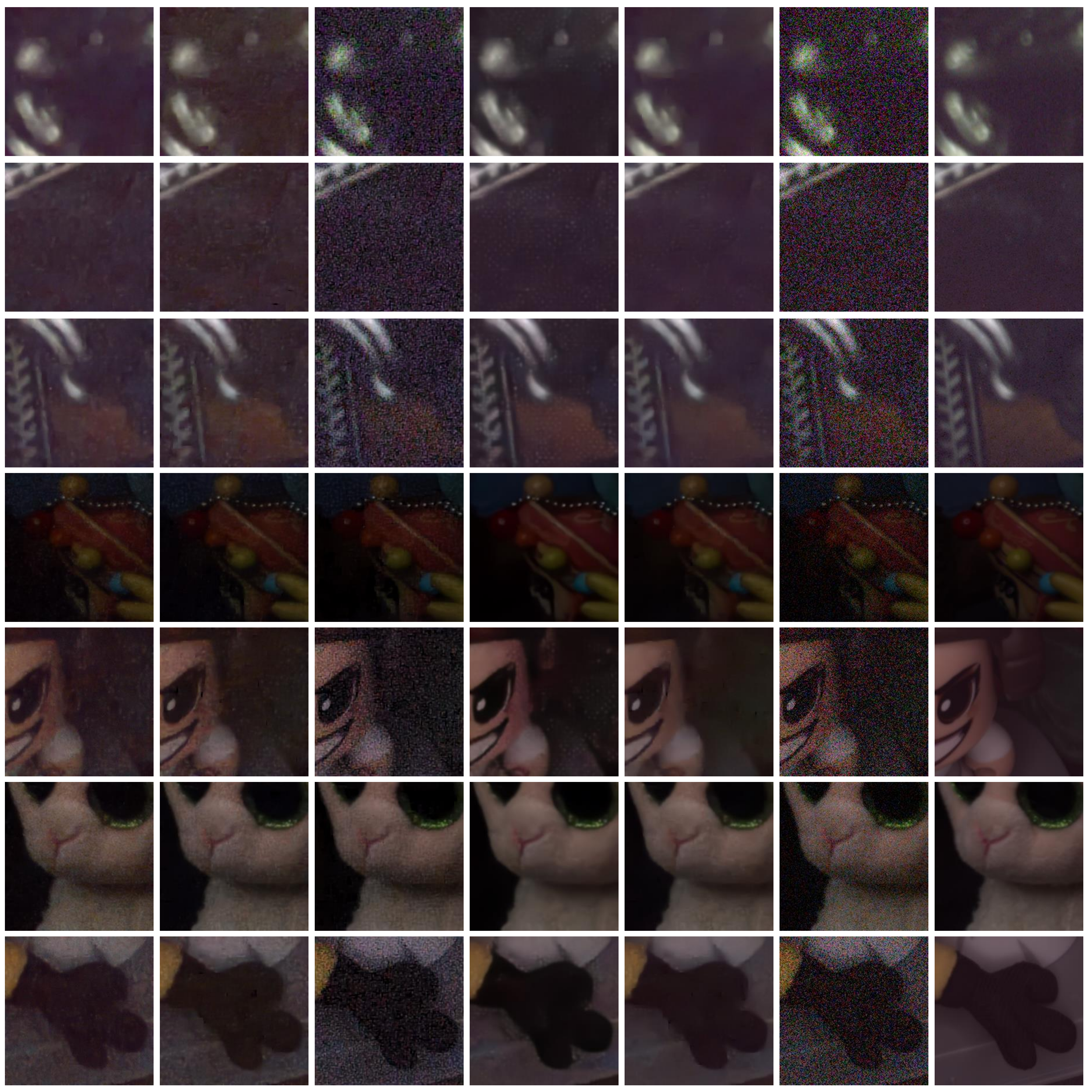}
  \caption{Visual results for real noise removal on the SIDD validation set (first three rows) and SIDD+ set (last four rows). \textbf{Left to Right}: RIDNet results trained by DANet~\cite{DANet}, GDANet~\cite{DANet}, C2N~\cite{C2N}, CycleISP~\cite{CycleISP}, NoiseTransfer (Ours), and Noisy and Clean image.}
  \label{fig_sidd_denoising}
\end{figure*}

\subsection{Discussion}
% {grdn,noiseflow,CA-NoiseGAN}: require metadata
% DANet: gaussian noise input // cannot control
% C2N: r map input // cannot know what desired noisy values like 
% PNGAN:
Our model has several advantages compared with existing noise generators.
% single network
\cite{syntGAN} trained 17 different generators to handle numerous camera models and ISO levels.
This solution could be straightforward to cover various noise distributions, but training multiple generators for each different noise lacks practicality.
% metadata
To compensate this, image metadata of a noisy image can be exploited to sample a specific noise type \cite{grdn,noiseflow,CA-NoiseGAN}.
However, such external information is not always available in the real-world.
% dependency on pre-trained network
PNGAN \cite{PNGAN} requires pre-trained networks for training.
Specifically, it uses a camera pipeline modeling network \cite{CycleISP} to generate a noisy image that is further refined by generator.
It also employs a pre-trained denoising network as the regularizer.
This strategy makes the generated noisy image distribution dependent on pre-trained networks, which may not be suitable in several cases.
% desired noise sample
Although C2N~\cite{C2N} takes random vector that determines the property of synthesized noise, we do not know which value should be used for the random vector when a particular type of noise is required.
Compared with these models, our model can handle numerous noise distributions with a single generator, and does not need external resources, and synthesizes a desired noise by transferring the noise information from a reference noisy image.

\section{Experiments}

\subsection{Implementation Details}\label{imple_details}
We train our NoiseTransfer model by using various synthetic and real noisy images.
First, for real-world noise, we use the SIDD-Medium dataset~\cite{SIDD} following previous works \cite{noiseflow,CA-NoiseGAN,DANet,C2N}.
In this case, two different patches are randomly selected from the same noisy image to get $Y$ and $Y^{r}$.
For synthetic noise, we sample noise from Gaussian distribution ($\sigma \in [0,70]$), Poisson distribution ($\lambda \in [5,100]$), and the combined Poisson-Gaussian distribution ($\sigma \in [0,70]$ and $\lambda \in [5,100]$).
% In addition, we also use a realistic noise introduced in CBDNet~\cite{CBDNet}, which considers an in-camera image signal processing pipeline with 201 different camera response functions.
Then, we acquire synthetic noisy images by corrupting clean images in the DIV2K training set~\cite{DIV2K} and SIDD-Meidum set using the noise from these synthetic distributions.
We use 32 mini-batch of 96 $\times$ 96 patches for training.
Each mini-batch includes 16 patches from the SIDD-Medium set and 16 patches corrupted with synthetic noise distributions.
We apply data augmentation (flip and rotation) to diversify train images.
% The SIDD-Medium set \cite{SIDD} has limited training set, so we apply several data augmentation techniques besides flip and rotation; noisy image interpolation \cite{syntGAN}, and CutMix \cite{cutmix} to diversify train images.
The noise embedding vector (\ie, outcome by $D_{noise}$) has 128-dimension, the size of the queue is set to 4096, and temperature parameter $\tau$ is set to 0.1 \cite{SupCon}.
$\lambda_{noise}^{FM},\lambda_{gan}^{FM}$, and $\lambda_{recon}$ are equally set to 100.
We use Adam optimizer~\cite{kingma2015adam} with an learning rate of 1e-4, $\beta_{1}$ = 0.5, and $\beta_{2}$ = 0.99.
We also apply L2 regularization with regularization factor 1e-7.
Our discriminator and generator are updated 2,000 times during one epoch, and training for 200 epochs takes approximately a week on two Tesla V100 GPUs.
We provide more details including network configurations and additional experimental results in the supplement.
% Our code will be publicly available upon acceptance.

\subsection{Noisy Image Generation}\label{sec_real_generation}
We first measure the accuracy of the generated noisy images.
To do so, we use Average KL Divergence (AKLD) value~\cite{DANet} and Kolmogorov-Smirnov (KS) test value\footnote{Histograms are computed with 256 bins evenly distributed in [-256,256].}~\cite{syntGAN} for quantitative evaluation, and compare the results with DANet \cite{DANet}, GDANet \cite{DANet}, C2N \cite{C2N}, and CycleISP \cite{CycleISP} in \tabref{tab_real_kld}.
Note that, DANet trained only with the SIDD-Medium dataset outperforms GDANet \cite{DANet} trained with three different real-noise datasets (SIDD-Medium, Poly~\cite{xu2018real}, and RENOIR~\cite{anaya2018renoir}).
The results demonstrate that GDANet does not handle the specific noise better in the SIDD dataset than DANet.
CycleISP \cite{CycleISP} samples random noise considering specific camera settings; hence, it is unlikely that distribution of the randomly sampled noise matches to that of noise within a specific noisy image.
By contrast, our NoiseTransfer which is trained with multiple different noise models can deal with the specific noise by transferring the noise characteristics within the reference noisy image, because our model utilizes noisy images as the reference $\tilde{Y}$ as well as clean images.
This advantage allows our model to obtain the best performance among the compared models.
However, it is worth mentioning that better AKLD/KS values do not always imply higher denoising performance as will be described in \secref{sec_real_denoising}.
AKLD/KS values compute the distance of pixel value distributions of two images, thus, we cannot predict how realistic generated noisy image is with only those values.

\figref{fig_sidd_generation} presents visual comparisons of generated noisy images.
The visual results show that our NoiseTransfer can synthesize more realistic and desirable noise than other models which frequently generate unexpected patterns.
Note that CA-NoiseGAN \cite{CA-NoiseGAN} conducts noise generation with raw images\footnote{For visualization, the camera pipeline matlab code (https://github.com/AbdoKamel/simple-camera-pipeline) is used.}, 
and NoiseGAN~\cite{syntGAN} only covers four ISO levels (400 $\sim$ 3200), thus, we provide only visual comparisons with these approaches.%\footnote{Github repository of PNGAN~\cite{PNGAN} does not provide a pre-trained model for noise generation, and only very few result examples are available.}.

\begin{table}[hbt!]
\centering
% \caption{AKLD/KS test values on the SIDD validation and SIDD+ datasets. Lower values indicate better noise generation performance. The best values are highlighted in bold.}
\caption{AKLD/KS test values on the SIDD validation and SIDD+ datasets. The best values are highlighted in bold.}
\label{tab_real_kld}
\resizebox{0.40\textwidth}{!}{%
\begin{tabular}{c|cc|cc}
Dataset  & \multicolumn{2}{c|}{SIDD validation} & \multicolumn{2}{c}{SIDD+}            \\ \hline
Metric   & \multicolumn{1}{c|}{AKLD$\downarrow$}   & KS$\downarrow$     & \multicolumn{1}{c|}{AKLD$\downarrow$}   & KS$\downarrow$     \\ \hline
DANet    & \multicolumn{1}{c|}{0.2117} & 0.0732 & \multicolumn{1}{c|}{0.4144} & 0.1468 \\
GDANet   & \multicolumn{1}{c|}{0.2431} & 0.1079 & \multicolumn{1}{c|}{0.4744} & 0.2557 \\
C2N   & \multicolumn{1}{c|}{0.3138} & 0.1499 & \multicolumn{1}{c|}{0.2882} & \textbf{0.1330} \\
CycleISP & \multicolumn{1}{c|}{0.6881} & 0.1743 & \multicolumn{1}{c|}{0.9412} & 0.1743 \\
Ours     & \multicolumn{1}{c|}{\textbf{0.1655}} & \textbf{0.0617} & \multicolumn{1}{c|}{\textbf{0.2324}} & 0.1537
\end{tabular}%
}
\end{table}

\subsection{Real Noise Denoising}\label{sec_real_denoising}
% training with SIDD-Medium
To more accurately validate the quality of generated noisy images, we evaluate the applicability of our NoiseTransfer in real-world image denoising.
In this work, we choose lightweight yet effective RIDNet \cite{RIDNet} as a baseline denoising network.
For a fair comparison, all generative noise models are evaluated by measuring the denoising performance of RIDNet.
% Note that other advanced denoising networks can be utilized without loss of generality.
Following previous works \cite{noiseflow,CA-NoiseGAN,DANet,C2N}, we use images from SIDD~\cite{SIDD} for training and validation.
% Then, our NoiseTransfer renders a new noisy patch $\tilde{Y}$ with $X$ and $Y^r$, and we use the pair of $X$ and $\tilde{Y}$ to train RIDNet.
The ground-truth clean images and the corresponding generated noisy images are used to train RIDNet.
% We do not include the ground-truth noisy images in the dataset to train the denoiser when evaluating a generative noise model.
We do not include the ground-truth noisy images in the dataset when training the denoiser to evaluate generative noise models.
For our NoiseTransfer, we randomly select a clean image $X$ and choose another random noisy image as the reference $Y^r$ from the SIDD-Medium dataset to render a new noisy patch $\tilde{Y}$.

% evaluation trained RIDNet
We evaluate the real noise removal performance on the SIDD validation, SIDD+ \cite{SIDD+}, and DND benchmark \cite{DND} datasets.
In \tabref{tab_real_denoising}, we measure the denoising performance in terms of PSNR and SSIM values, and denoising results by RIDNet trained with generated noisy images from DANet, GDANet, C2N, CycleISP, and our NoiseTransfer are compared.
We also present the result when the ground-truth noisy images are used instead of generated images (RIDNet+GT).
% Our outstanding denoising results show that proposed NoiseTransfer generates desirable noisy images for the different real noise datasets, and demonstrates that ours is generally applicable for training the real-noise denoiser.
Note that C2N~\cite{C2N} got the best KS value on SIDD+ in \tabref{tab_real_kld}, but PSNR/SSIM values are lower than other models.
This result shows AKLD/KS values do not always hint at higher denoising performance as stated in \secref{sec_real_generation}.
Our outstanding denoising results show that RIDNet trained with generated noisy images by our NoiseTransfer is generally applicable for real-world denoising.
Notably, our method got comparable denoising performance with `RIDNet+GT', especially on SIDD+, and this result is not surprising in this field. For example, NoiseFlow \cite{noiseflow} achieved better performance when using generated images during training rather than the ground-truth real images for raw image denoising (refer to Table. 3 in \cite{noiseflow}). This is due to the small number of real GT samples in the training dataset.
\figref{fig_sidd_denoising} shows visual denoising results on the SIDD validation and SIDD+ datasets.
% , and \figref{fig_dnd_denoising} provides denoising results on the DND benchmark dataset.

\begin{figure}[hbt!]
\centering
\begin{huge}
    \begin{minipage}[c]{0.48\linewidth}
    \centering
        \includegraphics[width=\linewidth]{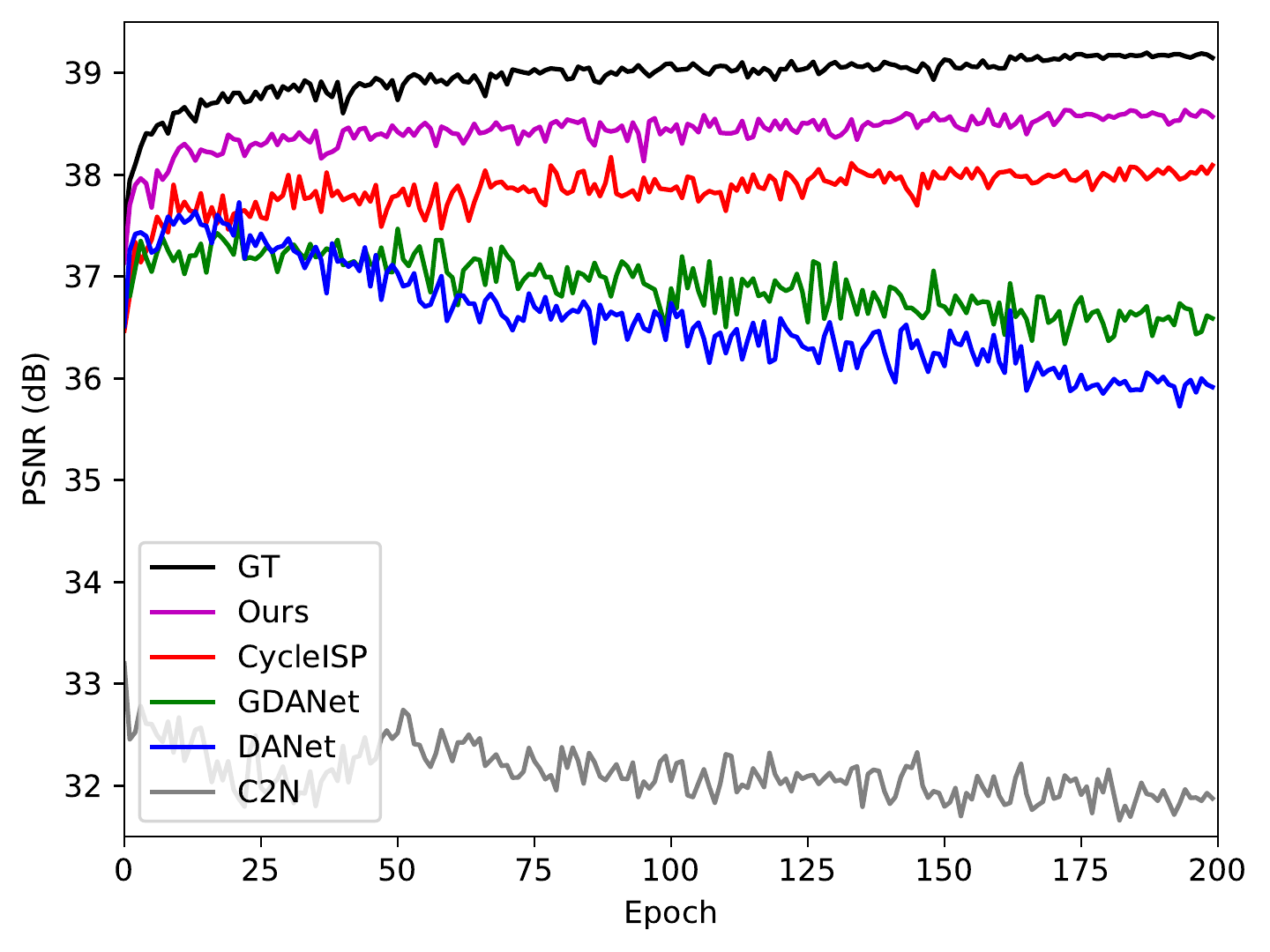}
    \end{minipage}
    \hfill
    \begin{minipage}[c]{0.48\linewidth}
    \centering
        \includegraphics[width=\linewidth]{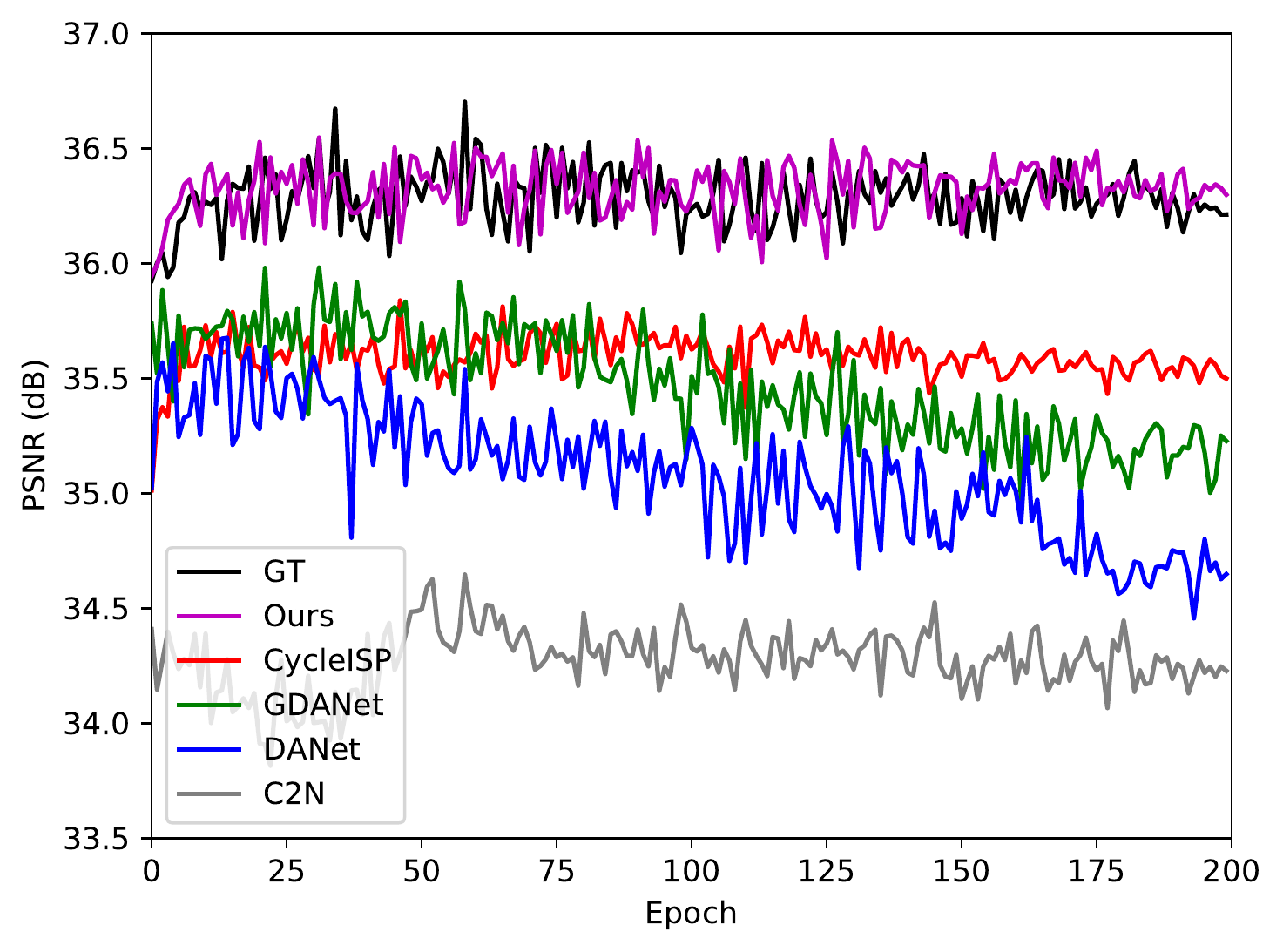}
    \end{minipage}
\caption{PSNR value changes during training on the SIDD validation set (Left) and SIDD+ (Right). Denoising performance trained by GT noisy images, NoiseTransfer (Ours), CycleISP, GDANet, DANet, and C2N are compared.}
\label{fig_sidd_psnr_curve}
\end{huge}
\end{figure}

Moreover, we plot changes of PSNR values by RIDNet during training on the SIDD validation and SIDD+ datasets in \figref{fig_sidd_psnr_curve}.
Particularly, we observe that when the RIDNet is trained with noisy images generated by either GDANet or DANet, the denoiser is overfitted after some iterations and PSNR values drops.
We believe this overfitting problem can be caused by unrealistic patterns that GDANet and DANet produce as shown in \figref{fig_sidd_generation}.
Note that CycleISP is not a generative model and instead injects synthetic realistic noise, so RIDNet trained with noisy images synthesized by CycleISP does not suffer from the overfitting problem.
However it provides limited performance because CycleISP considers predetermined shot/read noise factors for specific camera settings to inject random noise, which may not follow distribution of real noise.
In contrast, RIDNet trained with images by our NoiseTransfer results show promising denoising results on several datasets (more than 0.5dB compared with CycleISP on average).
% Surprisingly, ours shows almost similar performance on the SIDD+ compared to RIDNet trained with the ground-truth noisy images.

\begin{table}[hbt!]
\centering
\caption{Denoising results in terms of PSNR/SSIM values on the various real-world noise datasets. `GT' denotes the ground-truth noisy images. \textcolor{red}{Red} and \textcolor{blue}{Blue} denote the best and second values.}
\label{tab_real_denoising}
\resizebox{0.48\textwidth}{!}{%
\begin{tabular}{l|c|c|c}
Dataset  & SIDD validation & SIDD+          & DND            \\ \hline
RIDNet + DANet    & 35.91 / 0.8762  & 34.64 / 0.8600 & 37.54 / 0.9292 \\
RIDNet + GDANet   & 36.58 / 0.8851  & 35.22 / 0.8844 & 37.13 / 0.9345 \\
RIDNet + C2N   & 31.87 / 0.7274  & 34.22 / 0.8290 & 35.41 / 0.9032 \\
RIDNet + CycleISP & 38.09 / 0.9095  & 35.49 / 0.9006 & 38.95 / 0.9485 \\
RIDNet + NoiseTransfer     & \textcolor{blue}{38.57} / \textcolor{blue}{0.9112}  & \textcolor{red}{36.30} / \textcolor{red}{0.9095} & \textcolor{blue}{39.15} / \textcolor{blue}{0.9492} \\
RIDNet + GT       & \textcolor{red}{39.14} / \textcolor{red}{0.9155}  & \textcolor{blue}{36.21} / \textcolor{blue}{0.9057} & \textcolor{red}{39.35} / \textcolor{red}{0.9502}
\end{tabular}%
}
\end{table}

% \begin{figure}[hbt!]
%   \centering
%   \includegraphics[width=\linewidth]{figs/fig_DND_denoising.pdf}
%   \caption{Visual results of noise removal on the DND dataset. (a) DANet~\cite{DANet}. (b) GDANet~\cite{DANet}. (c) CycleISP~\cite{CycleISP}. (d) NoiseTransfer (Ours). Note that the clean image is not publicly available.}
%   \label{fig_dnd_denoising}
% \end{figure}

\begin{table*}[hbt!]
\centering
\caption{PSNR/SSIM results of synthetic noise removal on the BSDS500 dataset. `GT' denotes the ground-truth noisy images. `$\text{N2G}_{g}$' and `$\text{N2G}_{p}$' denote two independently trained networks for Gaussian and Poisson noise respectively. For a random noise level, we report an average of 10 trials. \textcolor{red}{Red} and \textcolor{blue}{Blue} denote the best and second values respectively.}
\label{tab_statistical_denoising}
\resizebox{\textwidth}{!}{%
\begin{tabular}{l|cc|cc|cc}
Noise Type  & \multicolumn{2}{c|}{Gaussian}                        & \multicolumn{2}{c|}{Poisson}                         & \multicolumn{2}{c}{Poisson-Gaussian}                 \\ \hline
Noise level & \multicolumn{1}{c|}{Fixed $(\sigma$ = 25)}          & Random $(\sigma \in [0,50])$          & \multicolumn{1}{c|}{Fixed ($\lambda$ = 30)}         & Random ($\lambda \in [5,50]$)         & \multicolumn{1}{c|}{Fixed ($\lambda$ = 30, $\sigma$ = 25)}       & Random ($\lambda \in [5,50]$, $\sigma \in [0,50]$)       \\ \hline
RIDNet + $\text{N2G}_{g}$         & \multicolumn{1}{c|}{31.21 / 0.8806} & \textcolor{blue}{32.82} / 0.8818 & \multicolumn{1}{c|}{30.28 / 0.8702} & 29.52 / 0.8464 & \multicolumn{1}{c|}{28.80 / 0.8276} & 28.15 / 0.8070 \\
RIDNet + $\text{N2G}_{p}$         & \multicolumn{1}{c|}{31.17 / 0.8792} & 32.74 / 0.8825 & \multicolumn{1}{c|}{30.31 / 0.8703} & 29.56 / 0.8477 & \multicolumn{1}{c|}{28.93 / 0.8301} & 28.32 / 0.8106 \\
RIDNet + NoiseTransfer        & \multicolumn{1}{c|}{\textcolor{blue}{31.27} / \textcolor{blue}{0.8834}} & 32.76 / \textcolor{blue}{0.8872} & \multicolumn{1}{c|}{\textcolor{blue}{30.58} / \textcolor{blue}{0.8735}} & \textcolor{blue}{29.95} / \textcolor{blue}{0.8530} & \multicolumn{1}{c|}{\textcolor{blue}{29.16} / \textcolor{blue}{0.8344}} & \textcolor{blue}{28.71} / \textcolor{blue}{0.8190} \\
RIDNet + GT         & \multicolumn{1}{c|}{\textcolor{red}{31.48} / \textcolor{red}{0.8868}} & \textcolor{red}{33.20} / \textcolor{red}{0.8913} & \multicolumn{1}{c|}{\textcolor{red}{30.74} / \textcolor{red}{0.8774}} & \textcolor{red}{30.20} / \textcolor{red}{0.8608} & \multicolumn{1}{c|}{\textcolor{red}{29.31} / \textcolor{red}{0.8375}} & \textcolor{red}{28.85} / \textcolor{red}{0.8212}
\end{tabular}%
}
\end{table*}

\subsection{Synthetic Noise Denoising}
\begin{figure}[hbt!]
  \centering
  \includegraphics[width=\linewidth]{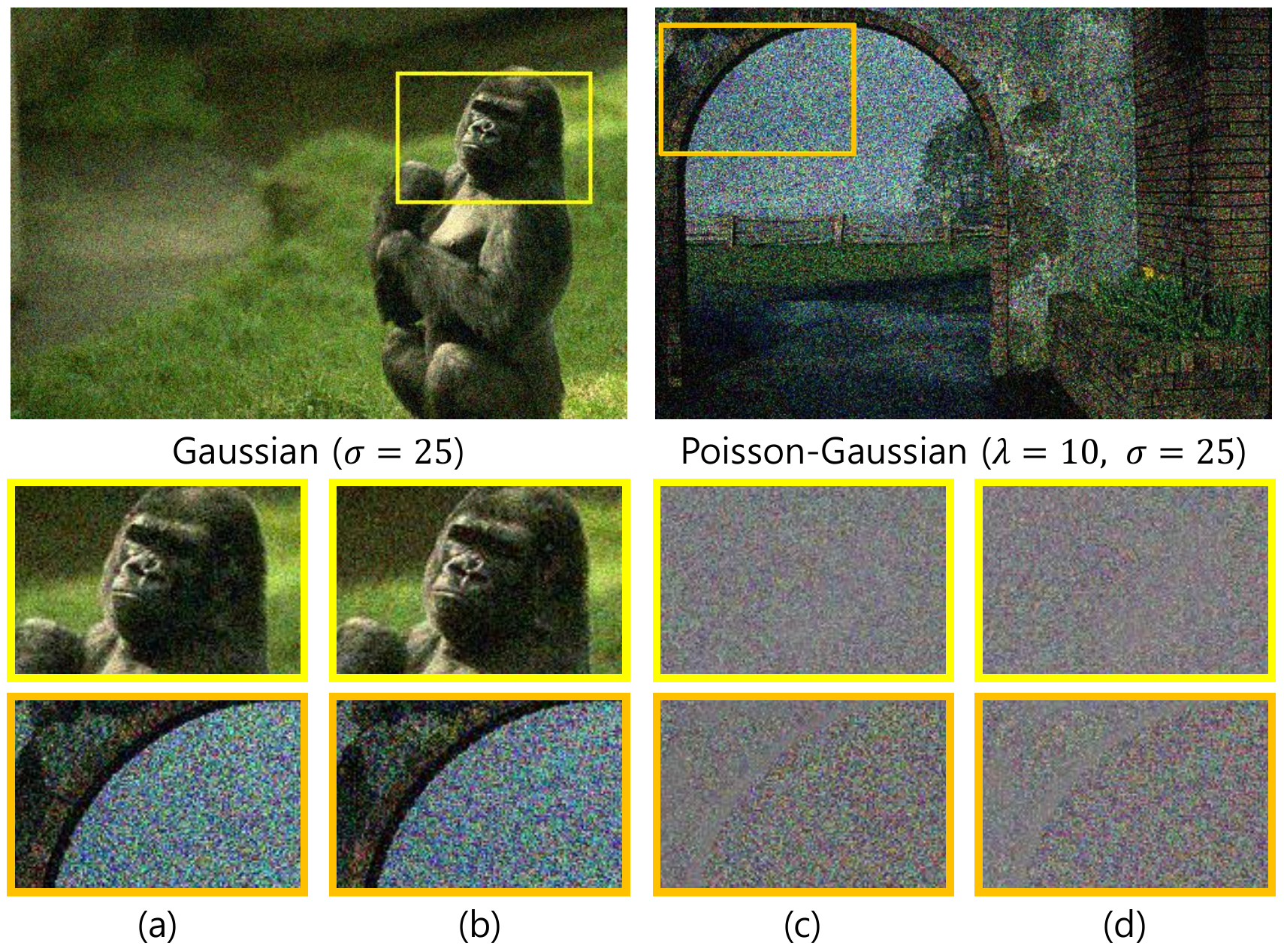}
  \caption{Examples of synthetic noise generation. (a) Ground-truth noisy image. (b) Generated noisy image by NoiseTransfer (Ours). (c)-(d) Noise added in (a) and (b), respectively.}
  \label{fig_statistical_generation}
\end{figure}

% 1.training
Finally, we evaluate the applicability of our NoiseTransfer in removing synthetic noise.
To do so, we first generate noisy images which include noise from known distributions using our NoiseTransfer.
% Specifically, we use randomly selected clean image in the DIV2K training set as $X$, and add one of the Gaussian noise ($\sigma \in [0,50]$), Poisson noise ($\lambda \in [5,50]$), and Poisson-Gaussian noise ($\sigma \in [0,50], \lambda \in [5,50]$) into another clean image for the reference noisy image $Y^r$.
Specifically, we use randomly selected clean image in the DIV2K training set as $X$, and add synthetic noise into another clean image for the reference noisy image $Y^r$.
We add one of Gaussian noise ($\sigma \in [0,50]$) and  Poisson noise ($\lambda \in [5,50]$) following N2G \cite{Noise2Grad}.
Additionally, we also add noise from Poisson-Gaussian distribution ($\sigma \in [0,50], \lambda \in [5,50]$) to confirm that our NoiseTransfer can generate diverse noises well.
The new noisy image $\tilde{Y}$ is synthesized with $X$ and $Y^r$, and RIDNet is trained with pairs of clean image $X$ and generated noisy image $\tilde{Y}$.
In \figref{fig_statistical_generation}, we present examples of our generated noisy images from known distributions, and we see that our model can synthesize signal-independent noise as well as signal-dependent one.

% 2.evaluation
To quantitatively measure the denoising performance, we use BSDS500 dataset \cite{amfm_pami2011} as testset and degrade images in the BSDS500 by adding noise from known distributions, and then put them into RIDNet as input.
Denoisng performance of RIDNet trained with noisy images by N2G \cite{Noise2Grad}, our NoiseTransfer, and the ground-truth noisy images are compared in \tabref{tab_statistical_denoising}.
Note that N2G does not generate noise, but instead extracts noise by denoising the input noisy image.
It also adopts a bernoulli random mask to destroy residual structure in the noise, and then the masked noise is used to corrupt other clean images.
In \tabref{tab_statistical_denoising}, N2G exhibits favorable performance when the noise type in N2G training matches the noise type in the test noisy image, but it reveals a slight performance drop against other types of noise.
In other words, $\text{N2G}_{g}$ shows slightly better denoising results for Gaussian noise removal and $\text{N2G}_{p}$ for Poisson noise removal.
This result implies that we may need to train multiple networks separately for each noise distribution (\eg, separated two networks for Gaussian and Poisson).
% By contrast, our NoiseTransfer can synthesize diverse types of noise with a single generator, and shows consistently better denoising performance for several types of noise.
By contrast, our NoiseTransfer shows consistently better denoising performance for several types of noise with a single generator, thus, our method does not require multiple generators independently trained for each noise type.

\subsection{Ablation Study}\label{exp_abl}
\begin{figure}[hbt!]
\centering
    \begin{minipage}[c]{0.48\linewidth}
    \centering
        \includegraphics[width=\linewidth]{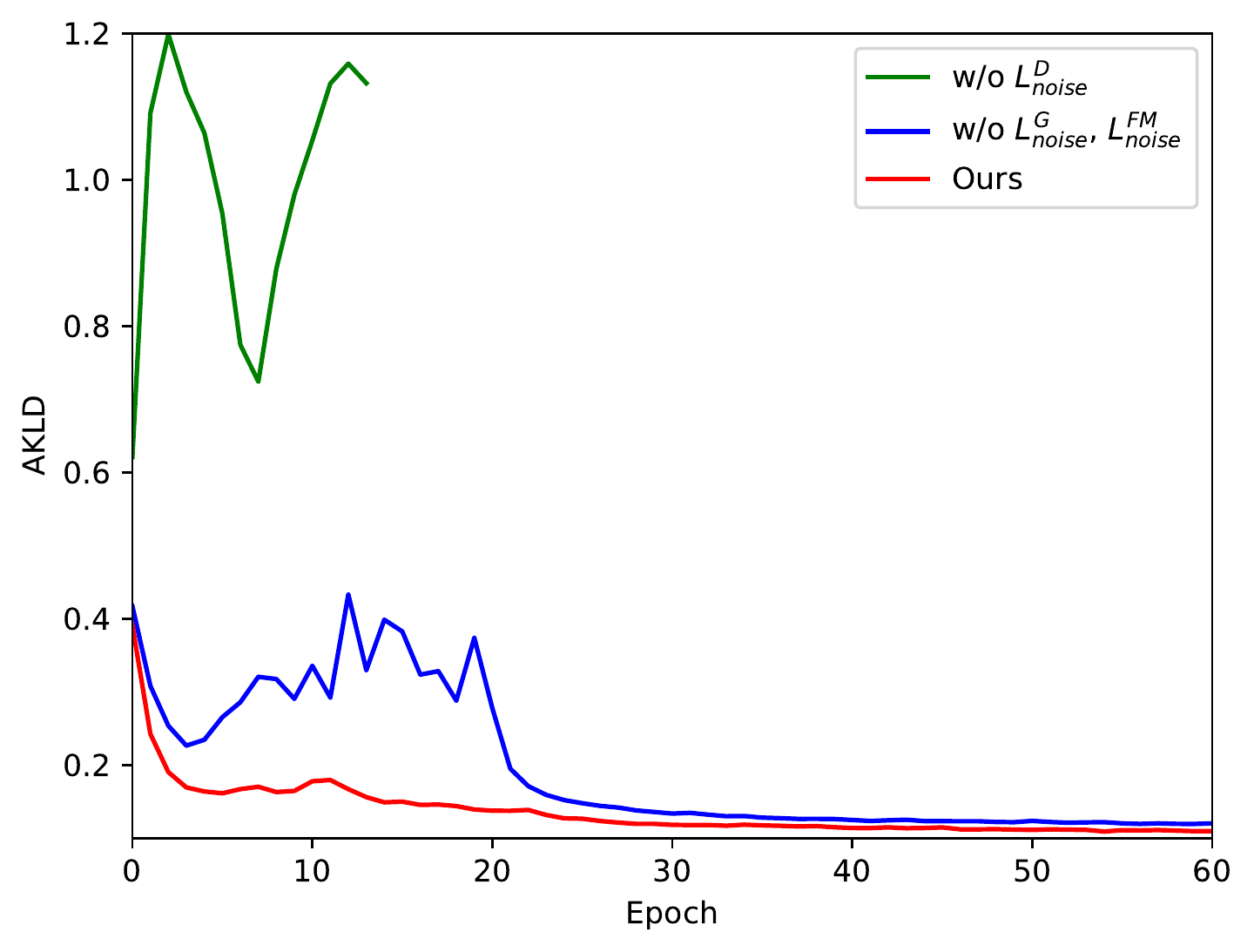}
        \subcaption{AKLD}
    \end{minipage}
    \hfill
    \begin{minipage}[c]{0.48\linewidth}
    \centering
        \includegraphics[width=\linewidth]{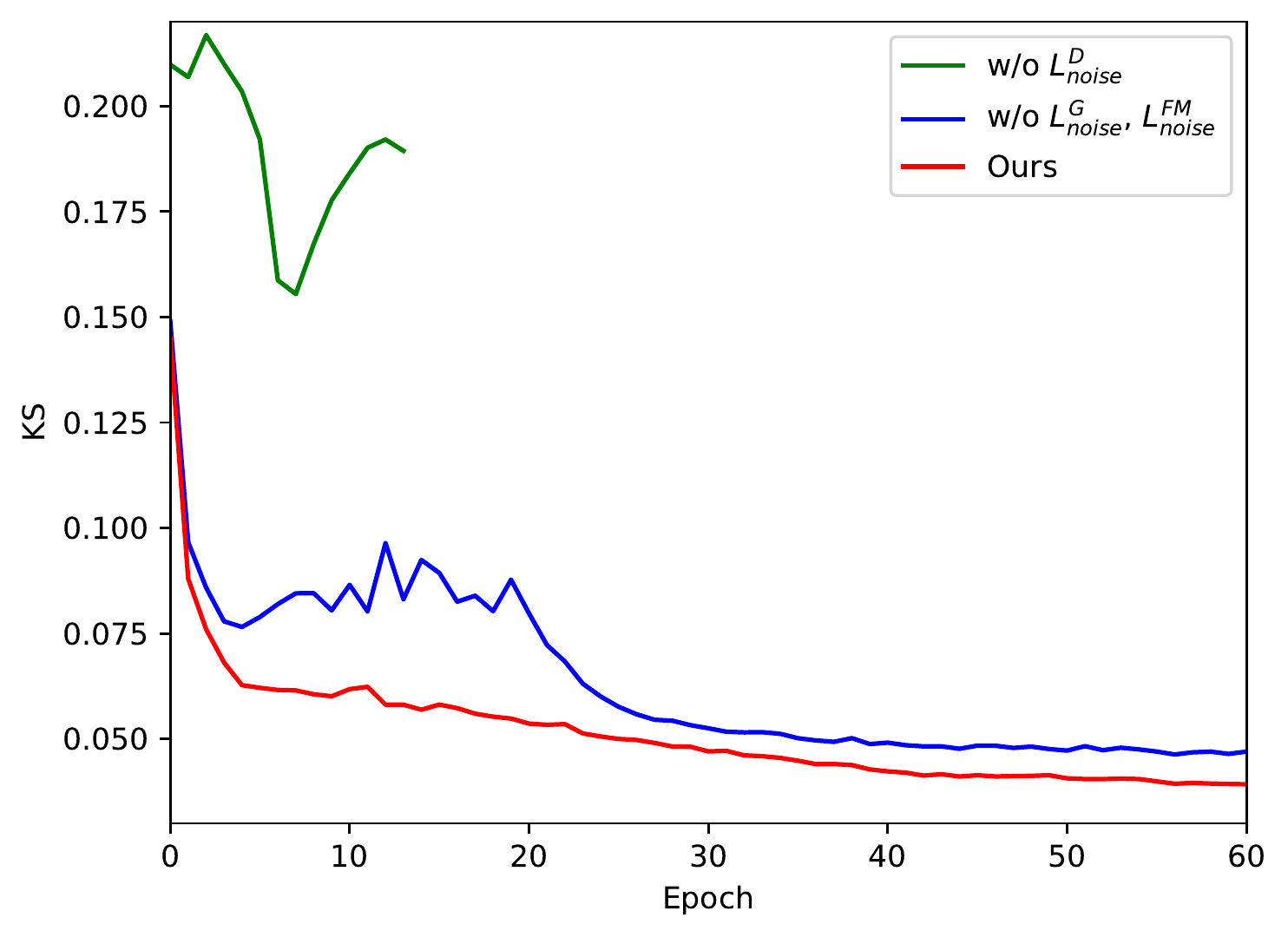}
        \subcaption{KS value}
    \end{minipage}
\caption{AKLD and KS test values for the first 60 training epochs. \textcolor{green}{(Green)} Trained without $L_{noise}^D$. This model diverged after 14 epochs. \textcolor{blue}{(Blue)} Trained without $L_{noise}^G$ and $L_{noise}^{FM}$. \textcolor{red}{(Red)} Our final NoiseTransfer.}
\label{fig_abl}
\end{figure}

In this work, we introduced contrastive losses for our NoiseTransfer.
Thus, we provide ablation study with and without using the additional contrastive losses during training.
We compare the accuracy of the generated noisy images in terms of AKLD and KS value.
\figref{fig_abl} obviously shows the effect of contrastive learning for image noise generation of our model\footnote{It compares the results for the first 60 training epochs, but is sufficient to confirm the effect of the proposed losses, allowing fair and efficient ablations.}.
First, without $L_{noise}^D$ which is the crux of our approach, we found the noise generation performance is very poor, and the model diverged after 14 epochs (Green).
This result demonstrates that learning distinguishable noise representation is crucial for our single generator to cover different kinds of noise distributions.
Next, with $L_{noise}^D$, we could see much better training results, but still have unstable performance early in training (Blue).
Finally, we can achieve more training stability and better performance when we explicitly guide our generator to synthesize a new noisy image with the same noise distribution to that of the $Y^r$ with additional losses $L_{noise}^G$ and $L_{noise}^{FM}$ (Red).

% \begin{figure}[hbt!]
% \centering
%     \begin{minipage}[c]{0.9\linewidth}
%     \centering
%         \includegraphics[width=\linewidth, height=0.5\linewidth]{figs/abl_AKLD.pdf}
%         \subcaption{AKLD}
%     \end{minipage}
%     \vfill
%     \begin{minipage}[c]{0.9\linewidth}
%     \centering
%         \includegraphics[width=\linewidth, height=0.5\linewidth]{figs/abl_KS.pdf}
%         \subcaption{KS value}
%     \end{minipage}
% \caption{AKLD and KS test values for the first 60 training epochs. \textcolor{green}{(Green)} Trained without $L_{noise}^D$. This model diverged after 14 epochs. \textcolor{blue}{(Blue)} Trained without $L_{noise}^G$ and $L_{noise}^{FM}$. \textcolor{red}{(Red)} Our final NoiseTransfer.}
% \label{fig_abl}
% \end{figure}

\section{Conclusion}
In this work, we proposed a novel noisy image generator trained with contrastive learning.
Different from existing works, our discriminator learns distinguishable noise representation for each different noise, which is the core of our method.
Thus, ours can extract noise characteristics from an input reference noisy image and generate new noisy images by transferring the specific noise to clean images.
%In other words, our discriminator extracts noise characteristics from an input reference noisy image.
This approach enables our generator to synthesize noisy images based on the noise information both in a paired or unpaired manner.
Consequently, our model can handle multiple noise distributions with a single generator.
Experiments demonstrate that the proposed generative noise model can produce more accurate noisy images than conventional methods and the applicability for image denoising.

%References are listed in alphabetic order by the surname of the first author, or the identifying word (e.g., in case of a website). Have
%all anonymized references at the beginning of the list.

%here would be your acknowledgement (if any) in the final accepted paper
\subsubsection{Acknowledgements}
This work was supported by Samsung Electronics Co., Ltd, and Samsung Research Funding Center of Samsung Electronics under Project Number SRFCIT1901-06.

%===========================================================

\bibliographystyle{splncs04}
\bibliography{egbib}

\end{document}

\title{NoiseTransfer: Image Noise Generation with Contrastive Embeddings \newline Supplementary material}
\titlerunning{NoiseTransfer}
\author{Seunghwan Lee \and
Tae Hyun Kim}
\authorrunning{Lee and Kim.}
\institute{Dept. of Computer Science, Hanyang University, Seoul, Korea
\email{\{seunghwanlee,taehyunkim\}@hanyang.ac.kr}}
\maketitle              % typeset the header of the contribution
%

%===========================================================

\section{Network Configurations} % fig

\subsection{Generator}
\begin{figure*}[h]
\centering

\includegraphics[width=0.9\linewidth]{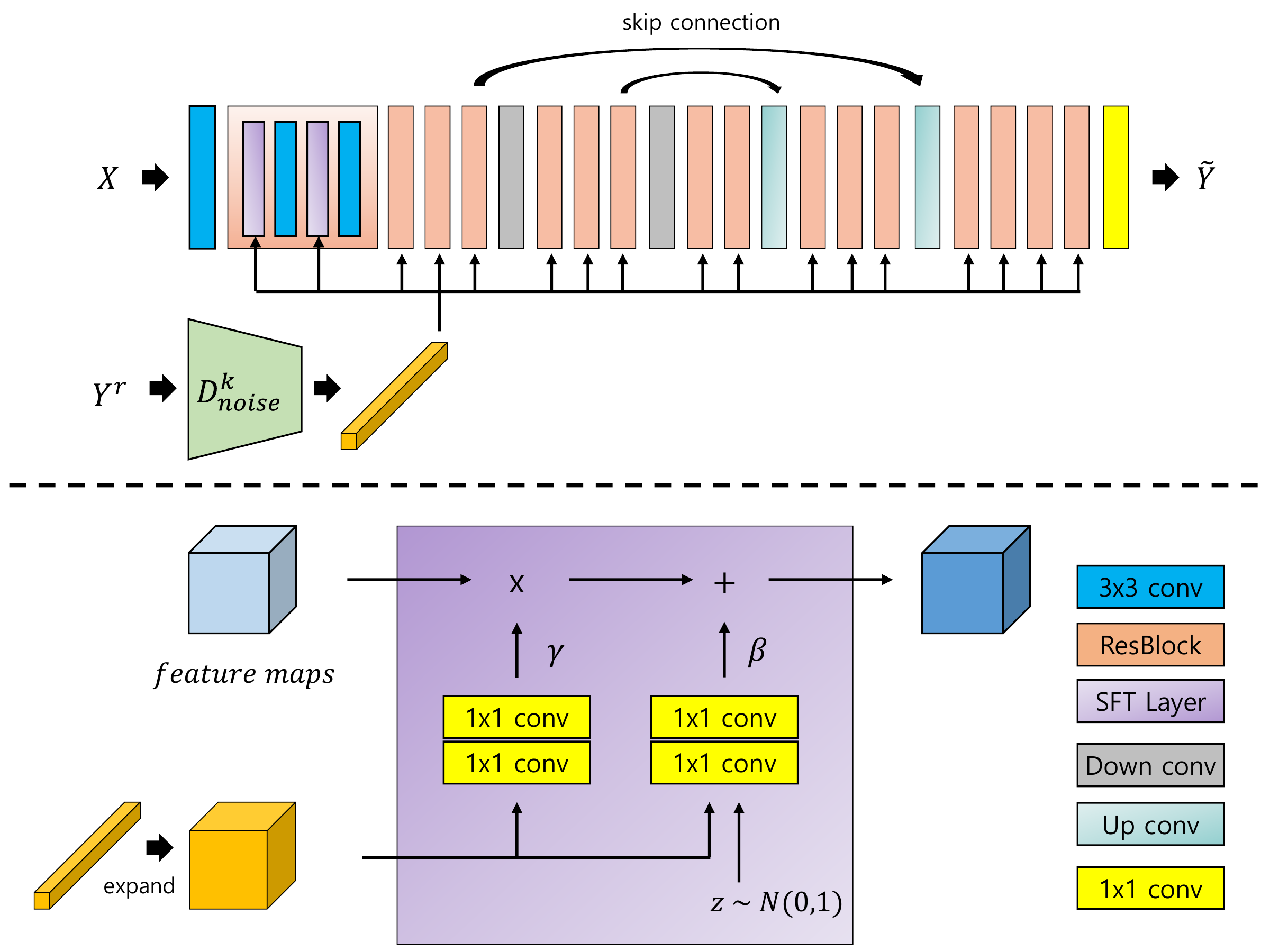}
\caption{Architecture of our U-Net based generator. The noise embeddings are expanded to match the shape of input feature maps in SFT layer \cite{SFTSR}. Gaussian random noise $z$ is concatenated with the noise embeddings for $\beta$ to generate spatially random noise.}
\label{fig_generator}

\end{figure*}

\figref{fig_generator} shows the architecture of our U-Net based generator.
We adopt the spatial feature transform (SFT) layer \cite{SFTSR} to efficiently translate features based on the noise embeddings.
Moreover, we add randomness in $\beta$ by introducing Gaussian random noise $z$ to generate differently realized noise for each forward.
We set the number of feature channels as 64, and we double the number of feature channels at each down-convolution.
LeakyRelu function is used for activation between convolution layers.

\subsection{Discriminator}
\begin{figure*}[h]
\centering

\includegraphics[width=0.9\linewidth]{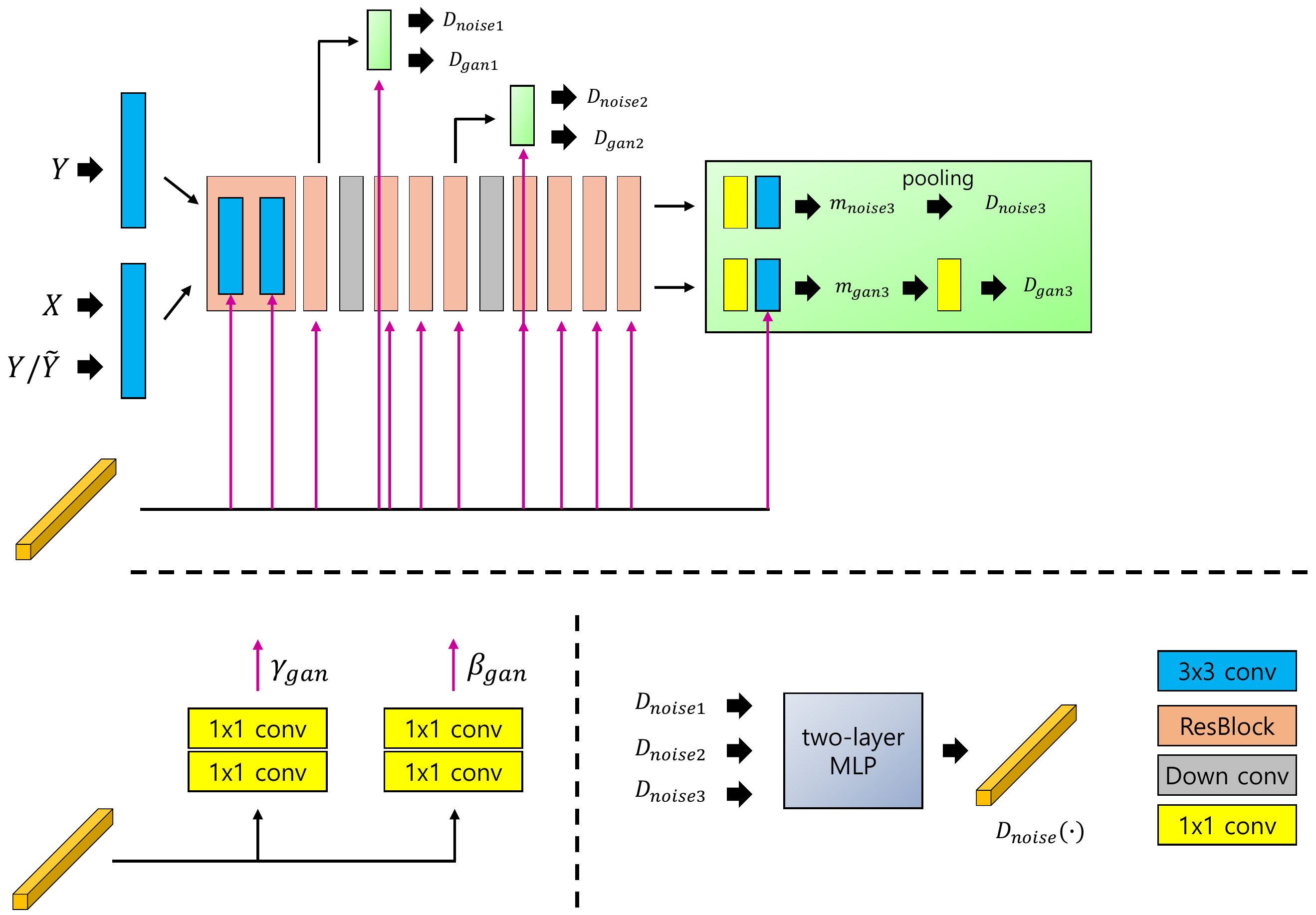}
\caption{Architecture of our discriminator. The discriminator has three levels of output with different receptive fields, and convolution weights in residual blocks are shared for $D_{noise}$ and $D_{gan}$. The scaling and shifting parameters for instance normalization in $D_{gan}$ are specially obtained from the noise embeddings, denoted by $\gamma_{gan}$ and $\beta_{gan}$ respectively. The noise embeddings are produced with the set of \{$D_{noise1}$, $D_{noise2}$, and $D_{noise3}$\} followed by MLP projection.}
\label{fig_discriminator}

\end{figure*}

\figref{fig_discriminator} shows the architecture of our discriminator.
Particularly, our discriminator has three levels of output with different receptive fields to benefit from various levels of contextual information from noisy image.
For outputs $D_{gan}(\cdot), m_{noise}(\cdot)$, and $m_{gan}(\cdot)$, the final outputs consist of the set of three levels of each output (\eg, $D_{gan}(\cdot)$ = \{$D_{gan1}(\cdot),D_{gan2(\cdot)}$, and $D_{gan3}(\cdot)$\}), and loss is computed for each set.
We adopt additional a two-layer multi-layer perceptron (MLP) to produce $D_{noise}(\cdot)$ from the set of \{$D_{noise1}(\cdot)$, $D_{noise2}(\cdot)$, and $D_{noise3}(\cdot)$\}, and ensure bounded values of noise embeddings by applying hyperbolic tangent function.
Instance normalization \cite{instance_norm} is utilized in the discriminator, however we derive the scaling and shifting parameters from the noise embeddings for $D_{gan}$.
Spectral normalization \cite{spectral_norm} is used for stable training for both generator and discriminator, and LeakyRelu function is used for activation.

% \section{Training Details}
% To train RIDNet \cite{RIDNet} with generated noisy images, we add L2 regularization loss with Adam optimizer to prevent overfitting (regularization factor $\lambda$ = 1e-5 for real noise removal and $\lambda$ = 1e-6 for synthetic noise removal).

\section{Limitation}
\begin{figure*}[h]
\centering

\includegraphics[width=0.9\linewidth]{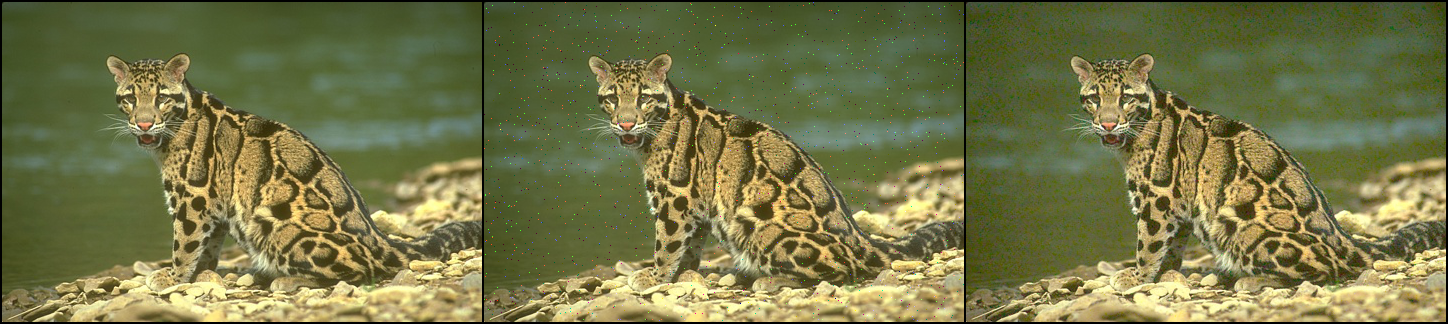}
\caption{Failure case of noise generation on the unseen noise type during training. The clean image is corrupted by salt and pepper noise. \textbf{Left to Right}: Clean, Noisy, Generated noisy image by our NoiseTransfer.}
\label{fig_failurecase}

\end{figure*}

We analyze the limitation of our model.
In fact, we do not guarantee that our model generates completely different types of noise that are not included in the training set.
We try to transfer and synthesize the unseen noise type during training, salt and pepper noise for example, as shown in \figref{fig_failurecase}.
Our model fails to generate a new noisy image with salt and pepper noise due to the mismatch of noise distributions of the training and inference, which is a common problem widely existing in data-driven deep learning methods.

\section{Visual Results}
\begin{figure*}[h]
\centering

\includegraphics[width=1.0\linewidth]{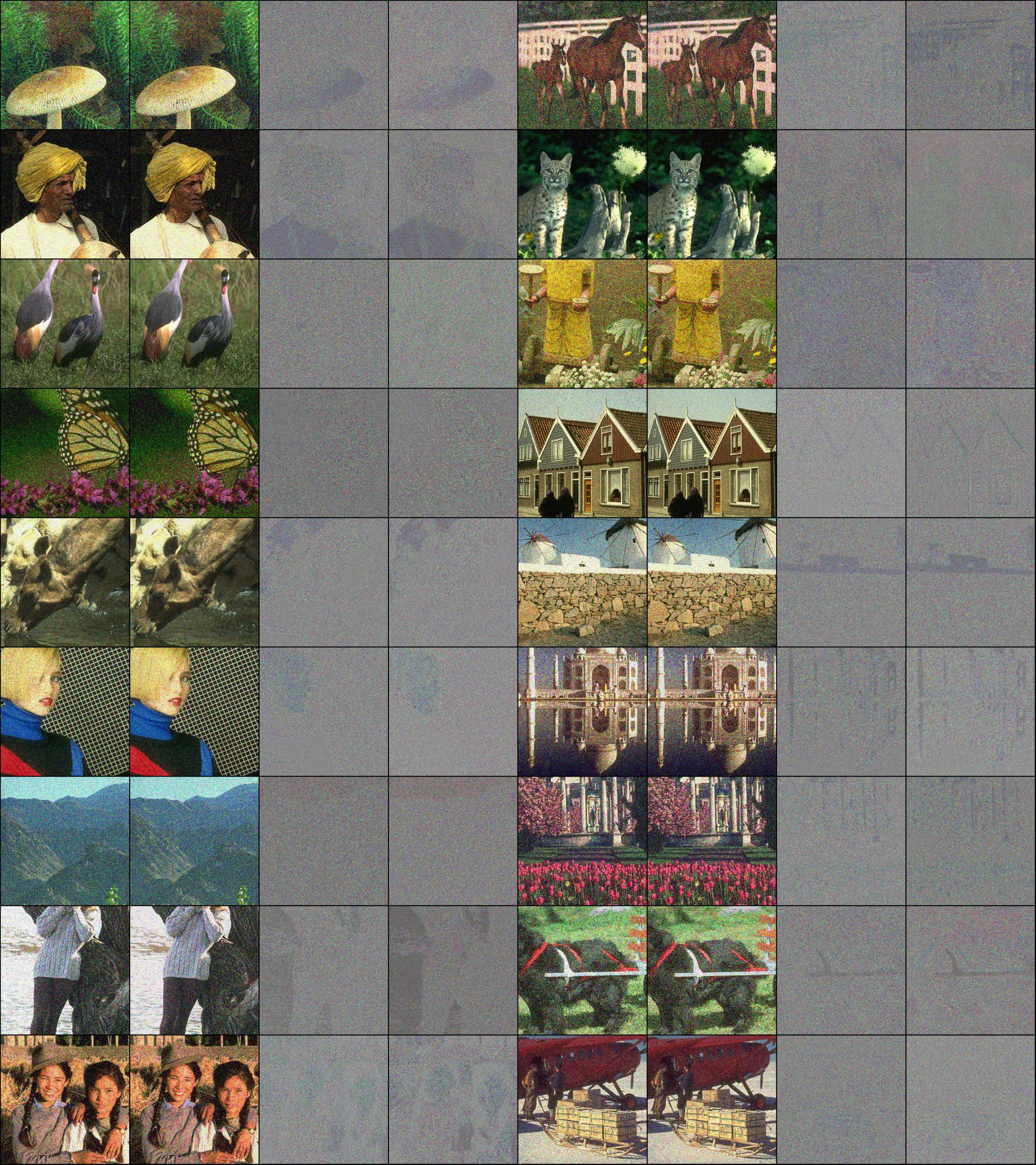}
\caption{Examples of synthetic noise generation. Gaussian (first three rows), Poisson (middle three rows), and Poisson-Gaussian noise (last three rows). \textbf{Left to Right for each instance}: Ground-truth noisy image, Generated noisy image by our NoiseTransfer, Ground-truth noise, Generated noise by NoiseTransfer (Ours).}
\label{fig_supp_synthetic_generation}

\end{figure*}

\begin{figure*}[h]
\centering

\includegraphics[width=1.0\linewidth]{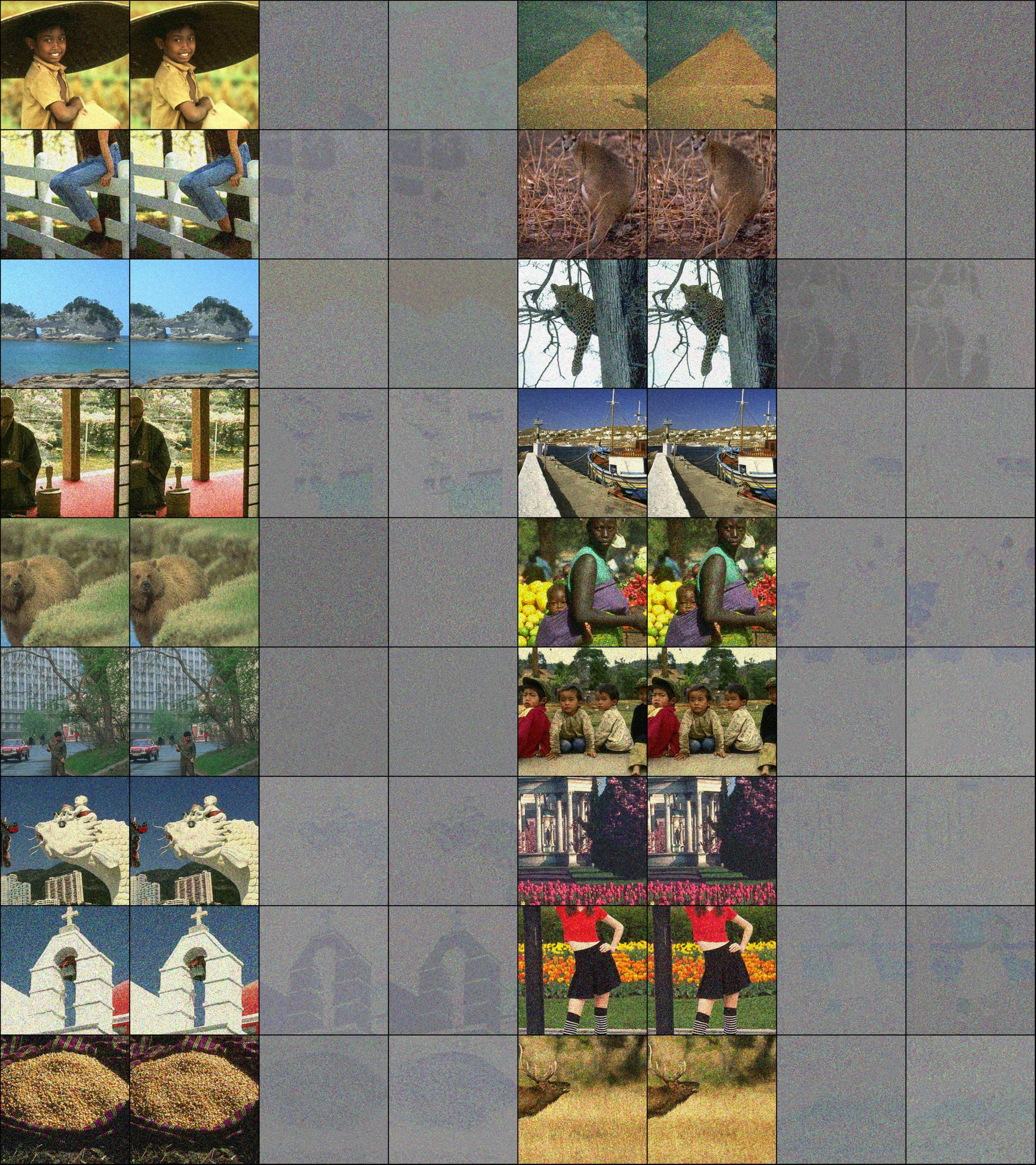}
\caption{Examples of synthetic noise generation. Gaussian (first three rows), Poisson (middle three rows), and Poisson-Gaussian noise (last three rows). \textbf{Left to Right for each instance}: Ground-truth noisy image, Generated noisy image by our NoiseTransfer, Ground-truth noise, Generated noise by NoiseTransfer (Ours).}
\label{fig_supp_synthetic_generation2}

\end{figure*}

\begin{figure*}[h]
\centering

\includegraphics[width=0.9\linewidth]{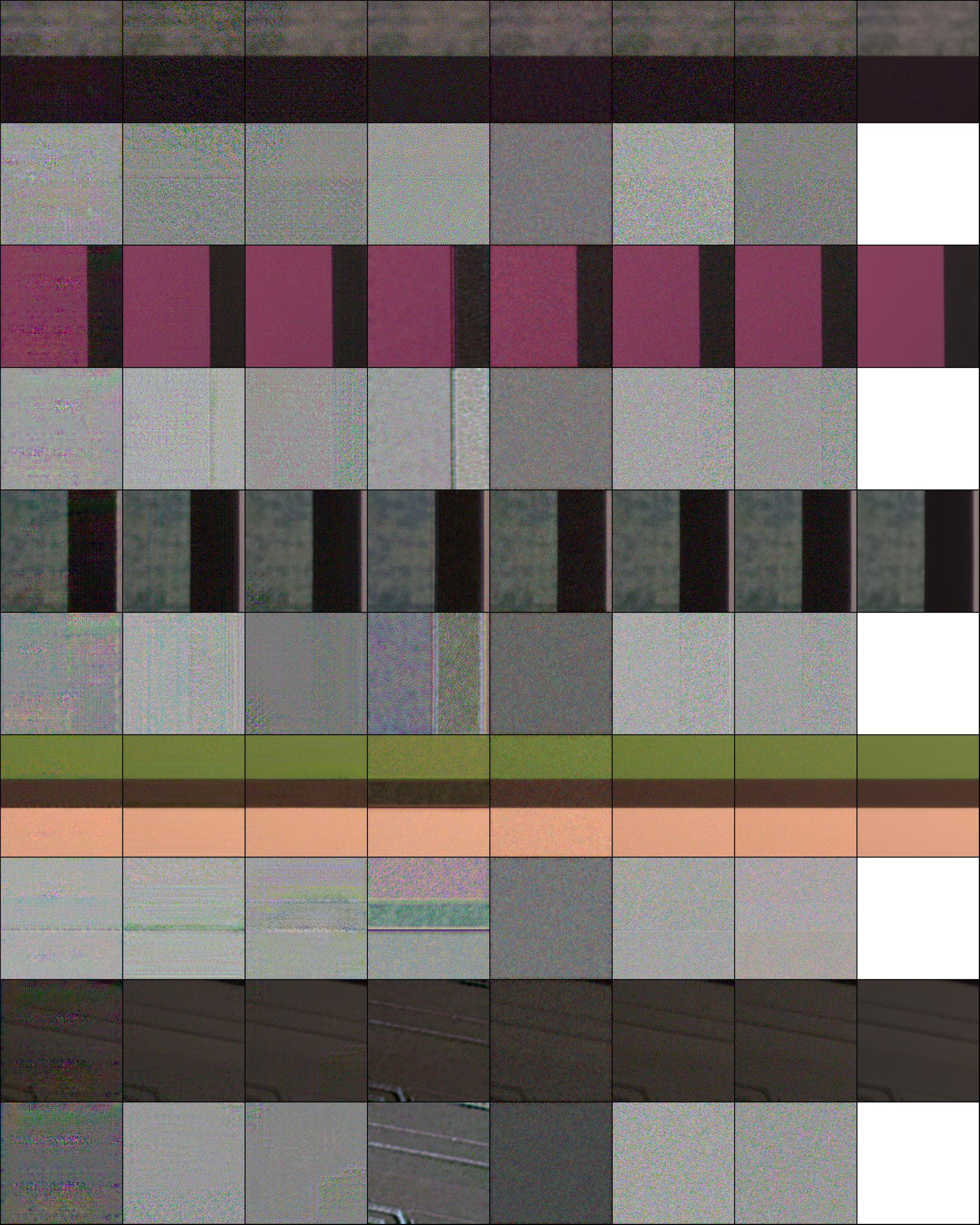}
\caption{Visual results of noise generation on the SIDD validation set. The corresponding noise is displayed below for each noisy image. \textbf{Left to Right}: CA-NoiseGAN~\cite{CA-NoiseGAN}, DANet~\cite{DANet}, GDANet~\cite{DANet}, NoiseGAN~\cite{syntGAN}, C2N~\cite{C2N}, NoiseTransfer (Ours), Noisy, Clean.}
\label{fig_supp_sidd_generation1}

\end{figure*}

\begin{figure*}[h]
\centering

\includegraphics[width=0.9\linewidth]{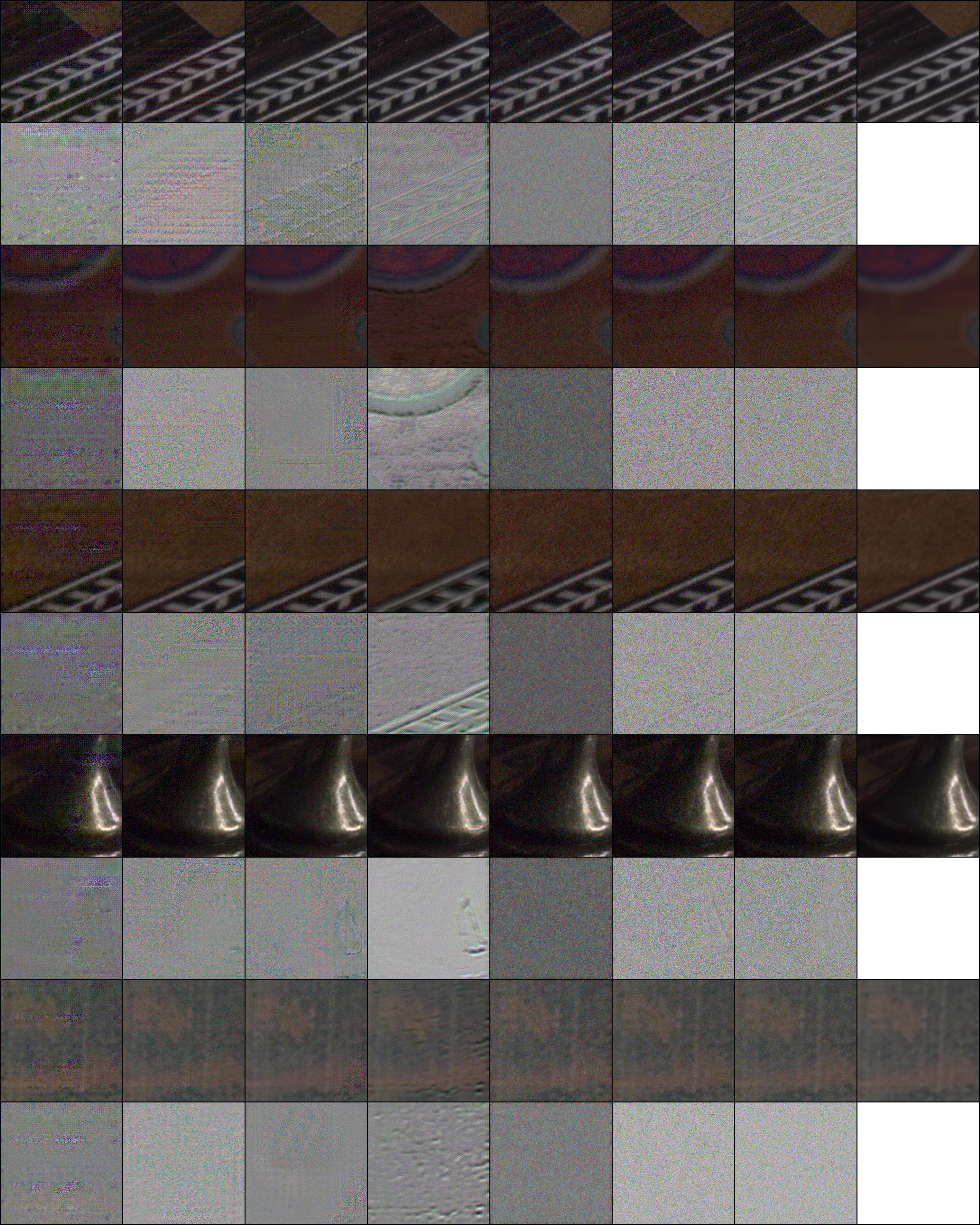}
\caption{Visual results of noise generation on the SIDD validation set. The corresponding noise is displayed below for each noisy image. \textbf{Left to Right}: CA-NoiseGAN~\cite{CA-NoiseGAN}, DANet~\cite{DANet}, GDANet~\cite{DANet}, NoiseGAN~\cite{syntGAN}, C2N~\cite{C2N}, NoiseTransfer (Ours), Noisy, Clean.}
\label{fig_supp_sidd_generation2}

\end{figure*}

\begin{figure*}[h]
\centering

\includegraphics[width=0.9\linewidth]{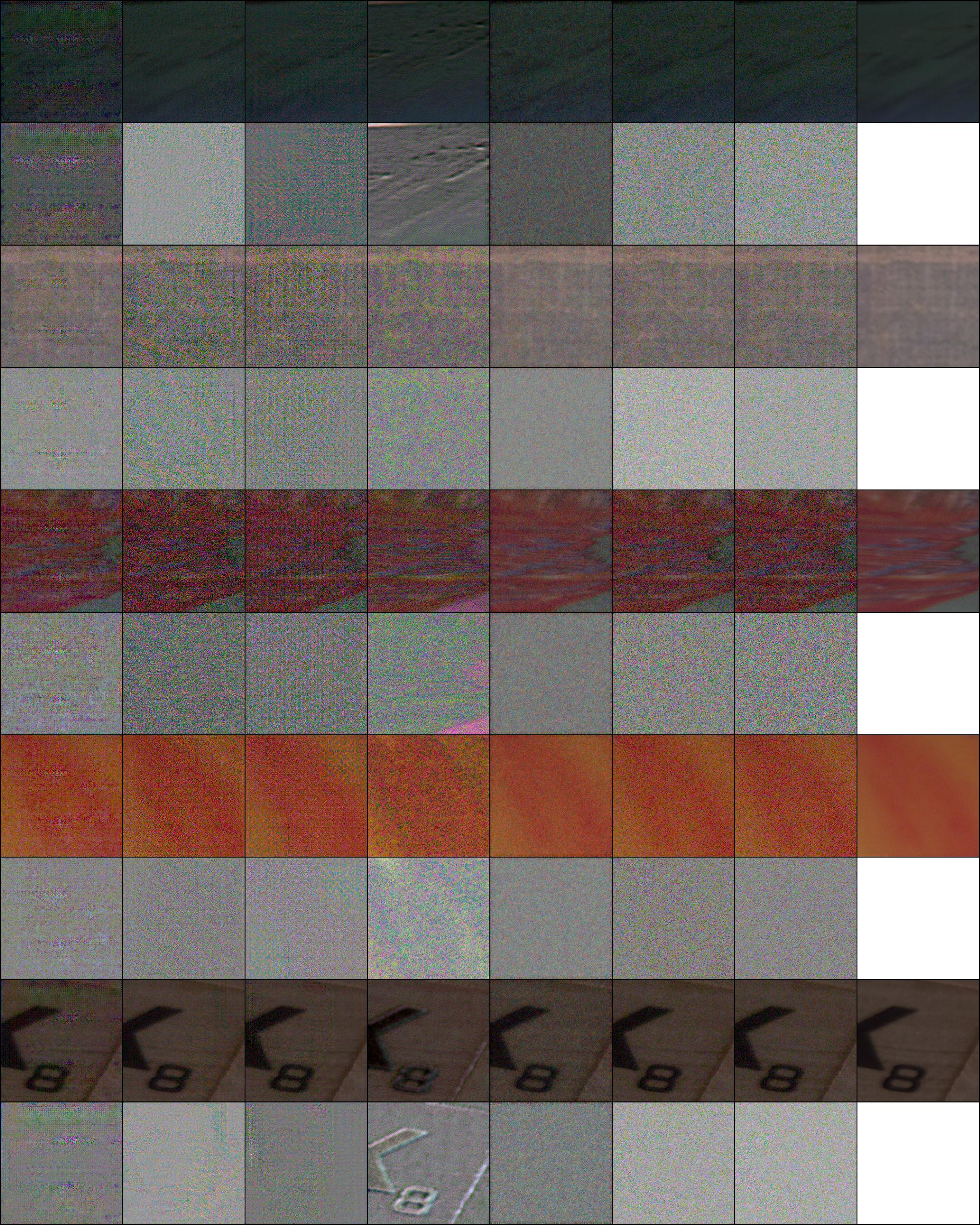}
\caption{Visual results of noise generation on the SIDD validation set. The corresponding noise is displayed below for each noisy image. \textbf{Left to Right}: CA-NoiseGAN~\cite{CA-NoiseGAN}, DANet~\cite{DANet}, GDANet~\cite{DANet}, NoiseGAN~\cite{syntGAN}, C2N~\cite{C2N}, NoiseTransfer (Ours), Noisy, Clean.}
\label{fig_supp_sidd_generation3}

\end{figure*}

\begin{figure*}[h]
\centering

\includegraphics[width=0.85\linewidth]{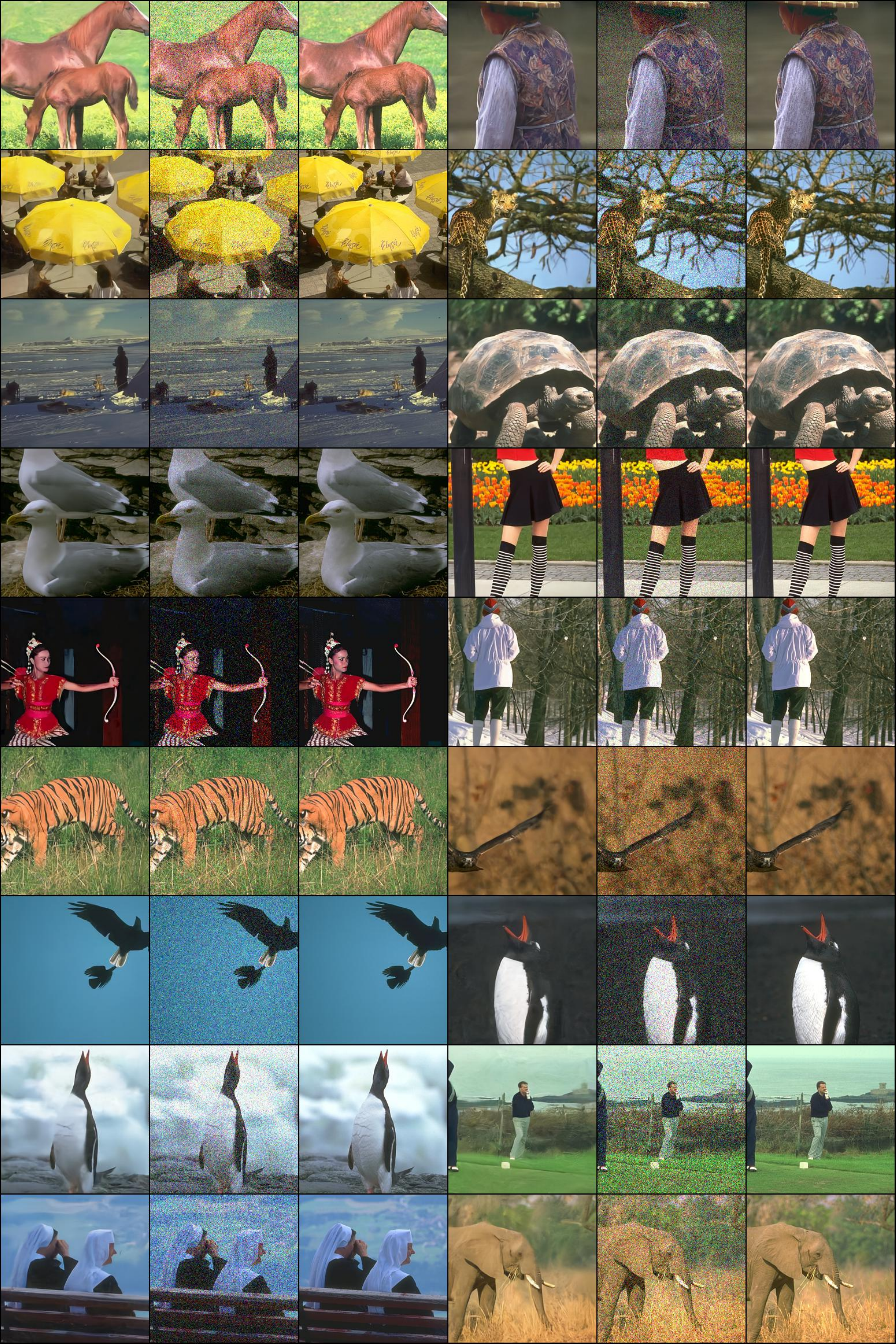}
\caption{Visual results for synthetic noise removal on the BSDS500. Gaussian (first three rows), Poisson (middle three rows), and Poisson-Gaussian noise (last three rows). \textbf{Left to Right for each instance}: RIDNet results trained by NoiseTransfer (Ours), and Noisy and Clean image.}
\label{fig_supp_synthetic_denoising}

\end{figure*}

\begin{figure*}[h]
\centering

\includegraphics[width=0.9\linewidth]{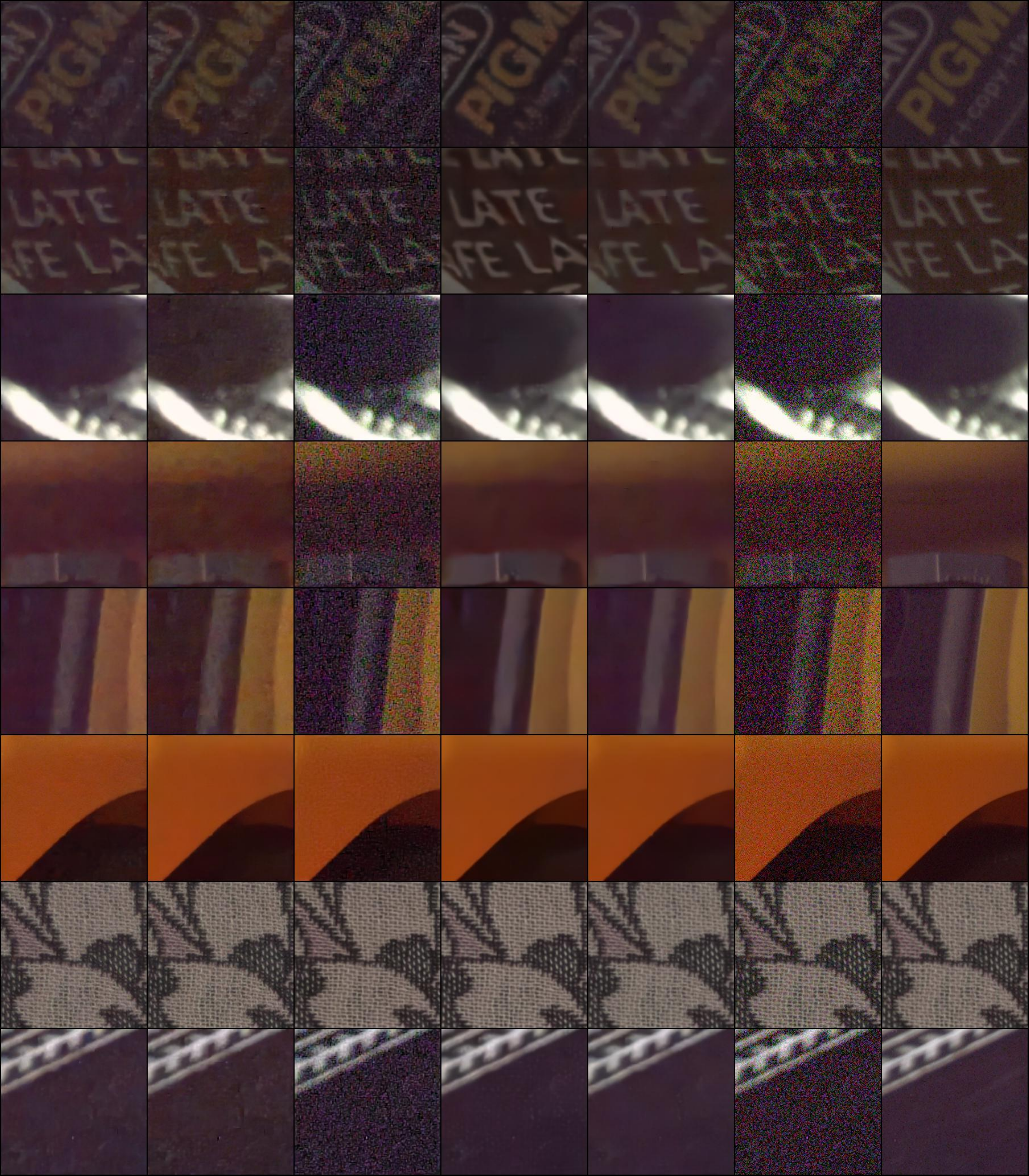}
\caption{Visual results for real noise removal on the SIDD validation set. \textbf{Left to Right}: RIDNet results trained by DANet~\cite{DANet}, GDANet~\cite{DANet}, C2N~\cite{C2N}, CycleISP~\cite{CycleISP}, and NoiseTransfer (Ours), and Noisy and Clean image.}
\label{fig_supp_sidd_denoising}

\end{figure*}

We provide more experimental results qualitatively.
\figref{fig_supp_synthetic_generation} and \figref{fig_supp_synthetic_generation2} show examples of synthetic noise generation for variable noise levels as stated in the manuscript (Sec. 4.1).
Visual comparisons for the real-world noise generation are presented from \figref{fig_supp_sidd_generation1} to \figref{fig_supp_sidd_generation3}.
\figref{fig_supp_synthetic_denoising} and \figref{fig_supp_sidd_denoising} show visual denoising results on the BSDS500 corrupted synthetically and on the SIDD validation set respectively.

% \begin{figure*}[h]
% \centering

% \includegraphics[width=0.9\linewidth]{figs/fig_supp_sidd_generation4.pdf}
% \caption{Visual results of noise generation on the SIDD validation set. The corresponding noise is displayed below for each noisy image. \textbf{Left to Right}: CA-NoiseGAN~\cite{CA-NoiseGAN}, DANet~\cite{DANet}, GDANet~\cite{DANet}, NoiseGAN~\cite{syntGAN}, NoiseTransfer (Ours), Noisy, Clean.}
% \label{fig_supp_sidd_generation4}

% \end{figure*}

%===========================================================

\bibliographystyle{splncs04}
\bibliography{egbib}